\documentclass[sigconf]{acmart}

\AtBeginDocument{%
  \providecommand\BibTeX{{%
    \normalfont B\kern-0.5em{\scshape i\kern-0.25em b}\kern-0.8em\TeX}}}

\setcopyright{acmlicensed}
\copyrightyear{2024}
\acmYear{2024}
\acmDOI{XXXXXXX.XXXXXXX}

\acmConference[MM'24]{Make sure to enter the correct
  conference title from your rights confirmation email}{October 28 - November 1,
  2024}{Melbourne, Australia.}
\acmISBN{978-1-4503-XXXX-X/18/06}

\usepackage[final]{pdfpages}
\begin{document}

\title{EALD-MLLM: Emotion Analysis in Long-sequential and De-identity videos with Multi-modal Large Language Model}


\author{Deng Li}
\affiliation{%
  \institution{Lappeenranta-Lahti University of Technology LUT}
  \city{Lappeenranta}
  \country{Finland}}
\email{deng.li@lut.fi}

\author{Xin Liu}
\affiliation{%
  \institution{Lappeenranta-Lahti University of Technology LUT}
  \city{Lappeenranta}
  \country{Finland}}
\email{xin.liu@lut.fi}

\author{Bohao Xing}
\affiliation{%
  \institution{Lappeenranta-Lahti University of
Technology LUT}
  \city{Lappeenranta}
  \country{Finland}}
\email{bohao.xing@lut.fi}

\author{Baiqiang Xia}
\affiliation{%
  \institution{Silo AI}
  \city{Helsinki}
  \country{Finland}}
\email{baiqiang.xia@silo.ai}

\author{Yuan Zong}
\affiliation{%
  \institution{Southeast Universit}
  \city{Nanjing}
  \country{China}}
\email{xhzongyuan@seu.edu.cn}

\author{Bihan Wen}
\affiliation{%
  \institution{Nanyang Technological University}
  \city{Singapore}
  \country{Singapore}}
\email{bihan.wen@ntu.edu.sg}

\author{Heikki Kälviäinen}
\affiliation{%
  \institution{Lappeenranta-Lahti University of Technology LUT}
  \city{Lappeenranta}
  \country{Finland}}
\email{heikki.kalviainen@lut.fi}

\renewcommand{\shortauthors}{Deng Li et al.}

\begin{abstract}
Emotion AI is the ability of computers to understand human emotional states. Existing works have achieved promising progress, but two limitations remain to be solved: 1) Previous studies have been more focused on short sequential video emotion analysis while overlooking long sequential video. However, the emotions in short sequential videos only reflect instantaneous emotions, which may be deliberately guided or hidden. In contrast, long sequential videos can reveal authentic emotions. 2) Previous studies commonly utilize various signals such as facial, speech, and even very sensitive biological signals (e.g., electrocardiogram). However, due to the increasing demand for privacy, it is becoming important to develop Emotion AI without relying on sensitive signals. To address the aforementioned limitations, in this paper, we construct a dataset for Emotion Analysis in Long-sequential and De-identity videos called \textit{EALD} by collecting and processing the sequences of athletes' post-match interviews. In addition to providing annotations of the overall emotional state of each video, we also provide Non-Facial Body Language (NFBL) annotations for each player. NFBL is an inner-driven emotional expression and can serve as an identity-free clue to understanding the emotional state. Moreover, we provide a simple but effective baseline for further research. More precisely, we propose a Multimodal Large Language Models (MLLMs) based solution for long-sequential emotion analysis called \texttt{EALD-MLLM} and evaluate it with de-identity signals (e.g., visual, speech, and NFBLs). Our experimental results demonstrate that: 1) MLLMs can achieve comparable, even better performance than the supervised single-modal models, even in a zero-shot scenario; 2) NFBL is an important cue in long sequential emotion analysis. EALD will be available on the open-source platform.
\end{abstract}





\maketitle

\section{Introduction}

Emotion analysis~\cite{koelstra2011deap, hakak2017emotion} is one of the most fundamental yet challenging tasks. For humans, emotions are present at all times, such as when people talking, thinking, and making decisions. Therefore, emotion analysis plays an important role in human-machine communication, such as in e-learning, social robotics, and healthcare~\cite{khare2023emotion,kolakowska2014emotion, cavallo2018emotion}. In the past decades, emotion analysis has attracted a lot of attention from the research community~\cite{nandwani2021review, li2020deep, el2011survey, ezzameli2023emotion, rahdari2019multimodal}. Various modalities of signals have been explored for emotion analysis. Biological signals such as electrocardiogram (ECG)~\cite{hsu2017automatic}, electroencephalogram (EEG)~\cite{li2022eeg}, and galvanic skin response~\cite{liu2016retracted} have been used for emotion analysis. Another research direction is static facial signals emotion analysis. For example, dataset EmotionNet~\cite{fabian2016emotionet} and AffectNet~\cite{mollahosseini2017affectnet} were proposed to identify emotions from facial images. Nowadays, video and multimodal (video + audio) emotion analysis has become a dominant research direction. AFEW~\cite{dhall2012collecting}, AFEW-VA~\cite{dhall2015video} and LIRIS-ACCEDE~\cite{baveye2015liris} provided emotional annotation from the movie clip. VAM~\cite{grimm2008vera} collected visual and speech data from the speech and interview. Similarly, MOSEI~\cite{zadeh2018multimodal} comprised 23,453 annotated video segments featuring 1,000 distinct speakers discussing 250 topics. The Youtube dataset~\cite{morency2011towards} included product reviews and opinion videos sourced from YouTube. SEWA~\cite{kossaifi2019sewa} recorded video and audio of subjects who watched adverts and provided detailed facial annotation. The summary of existing video and multimodal datasets for emotion analysis is present in Table.~\ref{tab:datasets}.

\begin{table*}[htbp]
    \centering
    \small
    \caption{Comparison of the different video and multimodal emotion understanding datasets.}
    \begin{tabular}{l|rrrrr}\hline\hline
         Dataset&Modality&Duration in total (hours)&Duration in avg. (mins)&NFBL annota.?&Identity free?  \\\hline
         HUMAINE~\cite{douglas2007humaine}&V+A+L&4.18 &5.02&\checkmark&$\times$\\
         VAM~\cite{grimm2008vera}&V+A&12.00&1.449&$\times$&$\times$\\
         IEMOCAP~\cite{busso2008iemocap}&V+A+L&11.46&<1&$\times$&$\times$\\
         Youtube~\cite{morency2011towards}&V+A+L&0.50&<1&$\times$&$\times$\\
         AFEW~\cite{dhall2012collecting}&V+A&2.46&<1&$\times$&$\times$\\
         AM-FED~\cite{mcduff2013affectiva}&V&3.33&<1&$\times$&$\times$\\
         AFEW-VA~\cite{dhall2015video}&V+A&0.66&<1&$\times$&$\times$\\
         LIRIS-ACCEDE~\cite{baveye2015liris}&V+L&-&<1&\checkmark&$\times$\\
         EMILYA~\cite{fourati2014emilya}&V+L&-&<1&\checkmark&$\times$\\
         SEWA~\cite{kossaifi2019sewa}&V+A&4.65&<1&$\times$&$\times$\\
         CMU-MOSEI~\cite{zadeh2018multimodal}&V+A+L&65.88&1.68&$\times$&$\times$\\
         iMiGUE~\cite{liu2021imigue}&V+L&34.78&5.81 &\checkmark&\checkmark\\
         EALD (Ours)&V+A+L&32.15&7.02&\checkmark&\checkmark\\\hline\hline
         \multicolumn{5}{l}{Note: V, A, and L denote video, audio, and language, respectively. }\\
    \end{tabular}
    \label{tab:datasets}
\end{table*}

\begin{figure}
    \centering
    \includegraphics[width=0.9\linewidth]{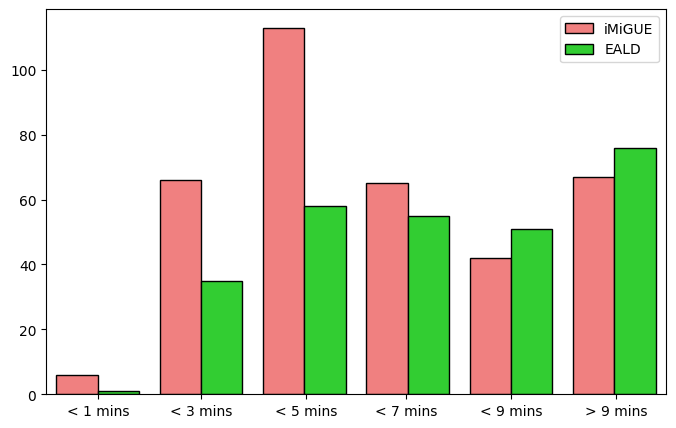}
    \caption{The comparison of the duration of videos of iMiGUE and \textit{EALD}. The X-axis denotes the length of the videos, and Y-axis denotes the number of videos. }
    \label{fig:imiguevseald}
\end{figure}

Previous datasets have successfully improved research in emotion analysis from various aspects. However, two major limitations remain to be addressed: 1) Regardless of the modality used, all of the above studies are highly correlated with sensitive biometric data. This is because biometric data (e.g., facial, speech, and biological signals) plays a critical role in various applications, such as telephone unlocks and mobile payment. While every coin has two sides, biometric information is so sensitive that it is particularly prone to being stolen, misused, and unauthorized tracking. As the risk of hacking and privacy violations increases, the protection of personal biometric data is receiving increasing attention; 2) Existing datasets are more focused on short sequential video emotion analysis while overlooking long sequential video emotion analysis. As shown in Table \ref{tab:datasets}, most datasets only provide videos that are shorter than one minute. The emotions depicted in short sequential videos merely capture fleeting moments, which could be intentionally directed or concealed. In contrast, long sequential videos have the capacity to unveil genuine emotions. For example, when an athlete who lost the game is interviewed after the game, although he or she may have expressed positive emotions in the middle of the interview, his or her overall emotion may still be negative. While iMiGUE~\cite{liu2021imigue} is close to our dataset, it lacks audio data. Moreover, as depicted in Fig. \ref{fig:imiguevseald}, the majority of videos in iMiGUE are shorter than 5 minutes, some even lasting only 3 minutes. Therefore, iMiGUE~\cite{liu2021imigue} may not be suitable for long-sequential emotion analysis.

\begin{figure}[htbp]
    \centering
    \tiny
    \setlength\tabcolsep{0.5pt}
    \begin{tabular}{ccc}
        \includegraphics[width=0.33\linewidth]{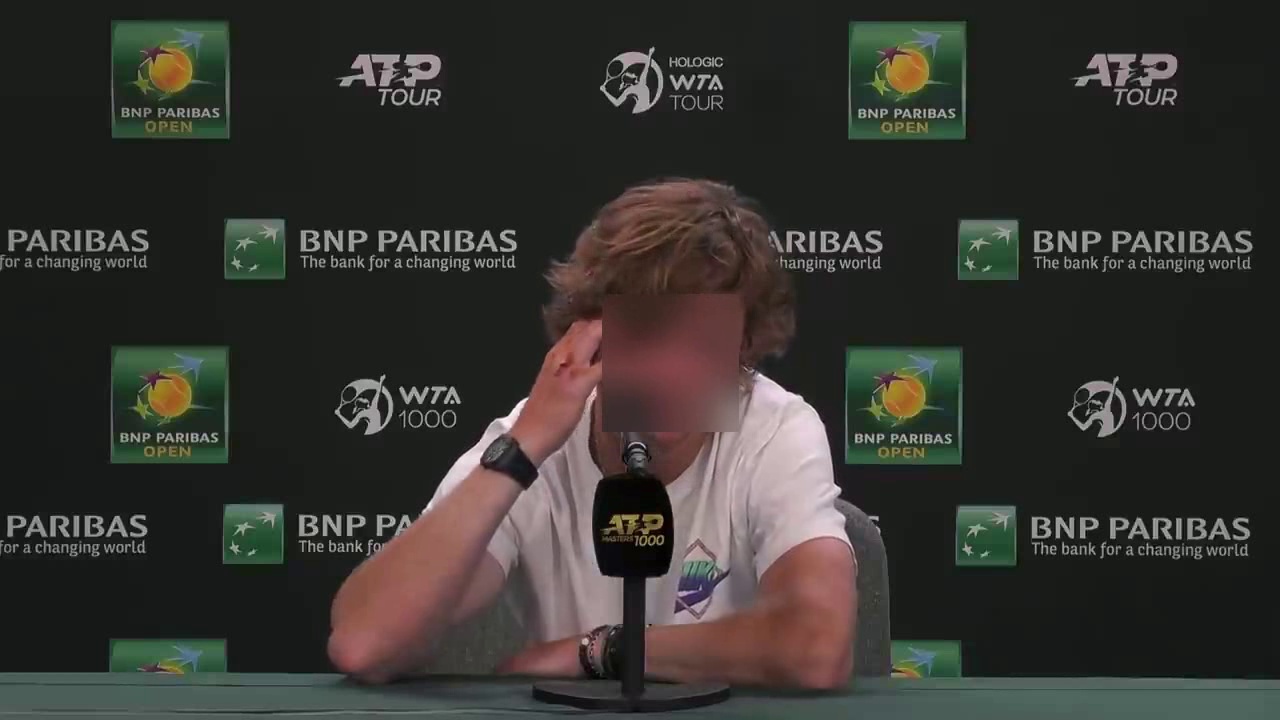}&\includegraphics[width=0.33\linewidth]{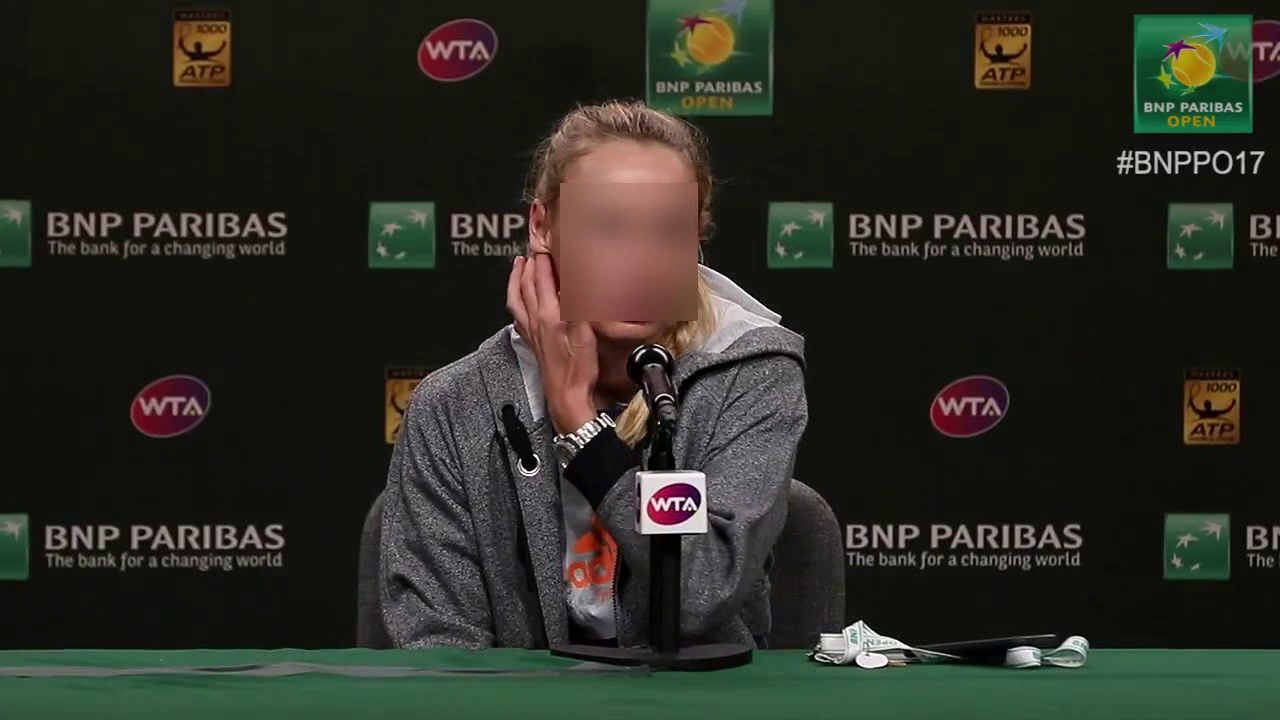}&\includegraphics[width=0.33\linewidth]{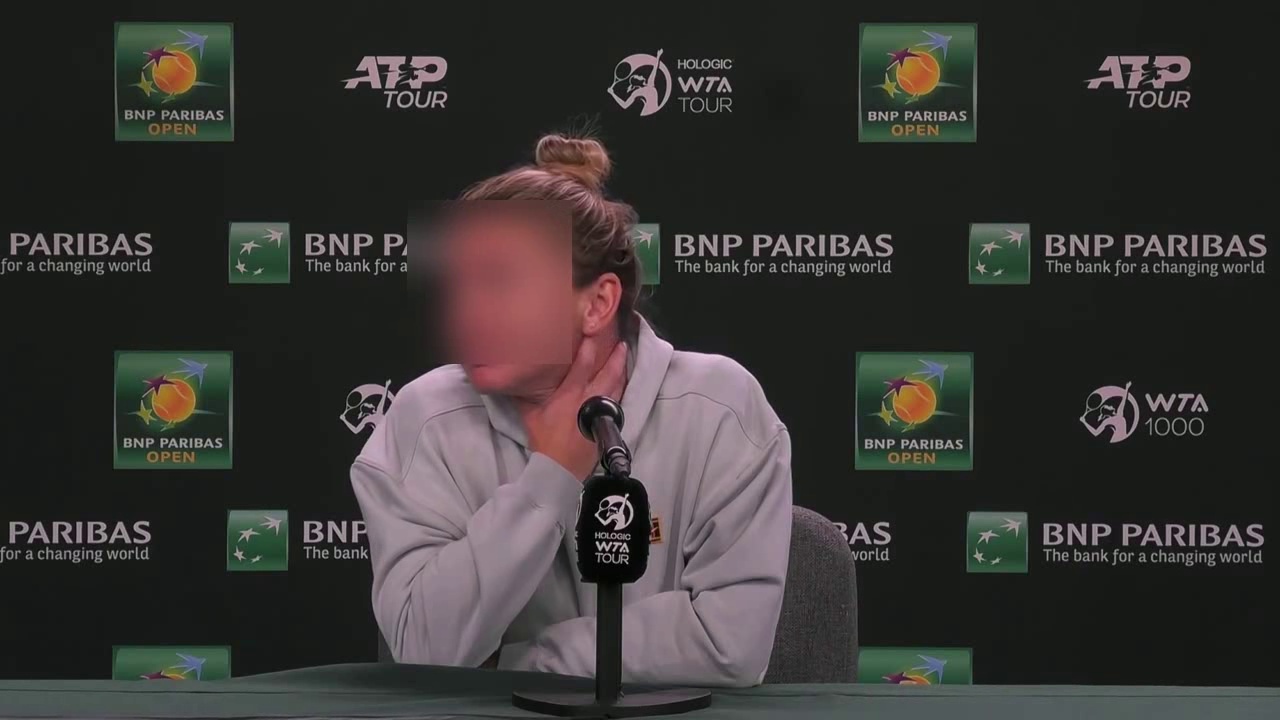}\\
        N3 Touching or scratching head&N8 Touching ears&N11 Touching or scratching neck\\
        \includegraphics[width=0.33\linewidth]{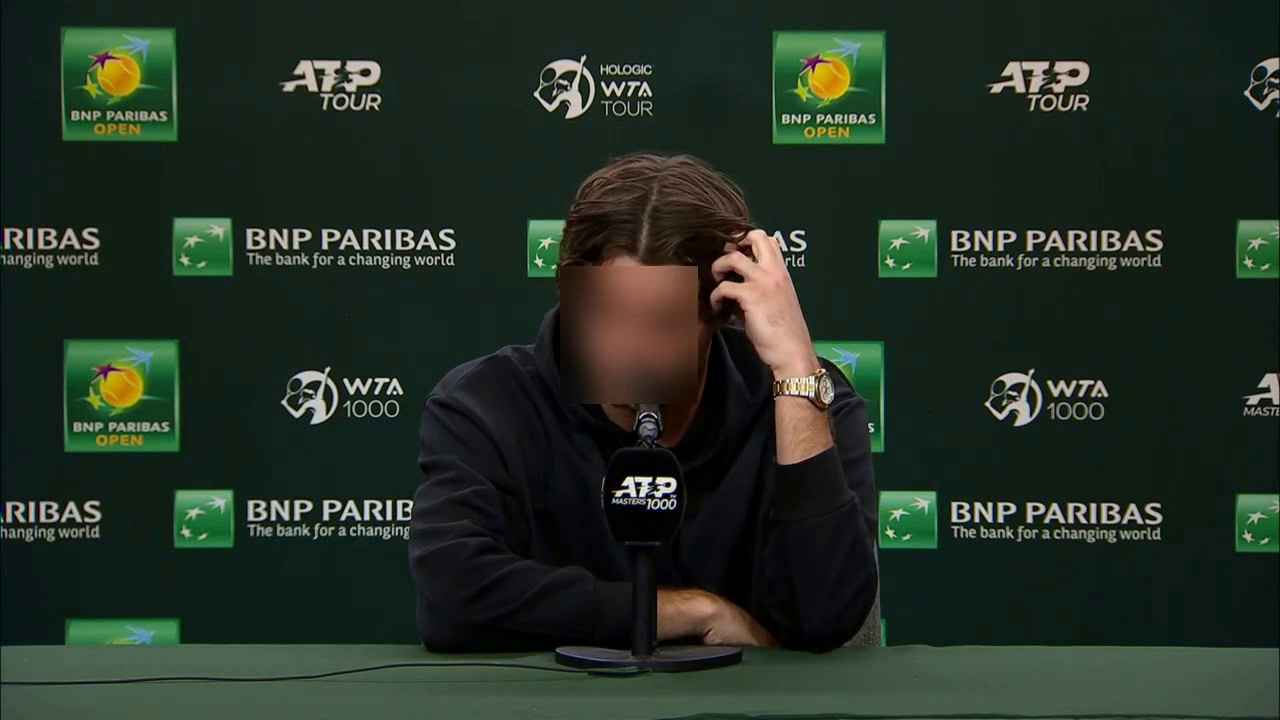}&\includegraphics[width=0.33\linewidth]{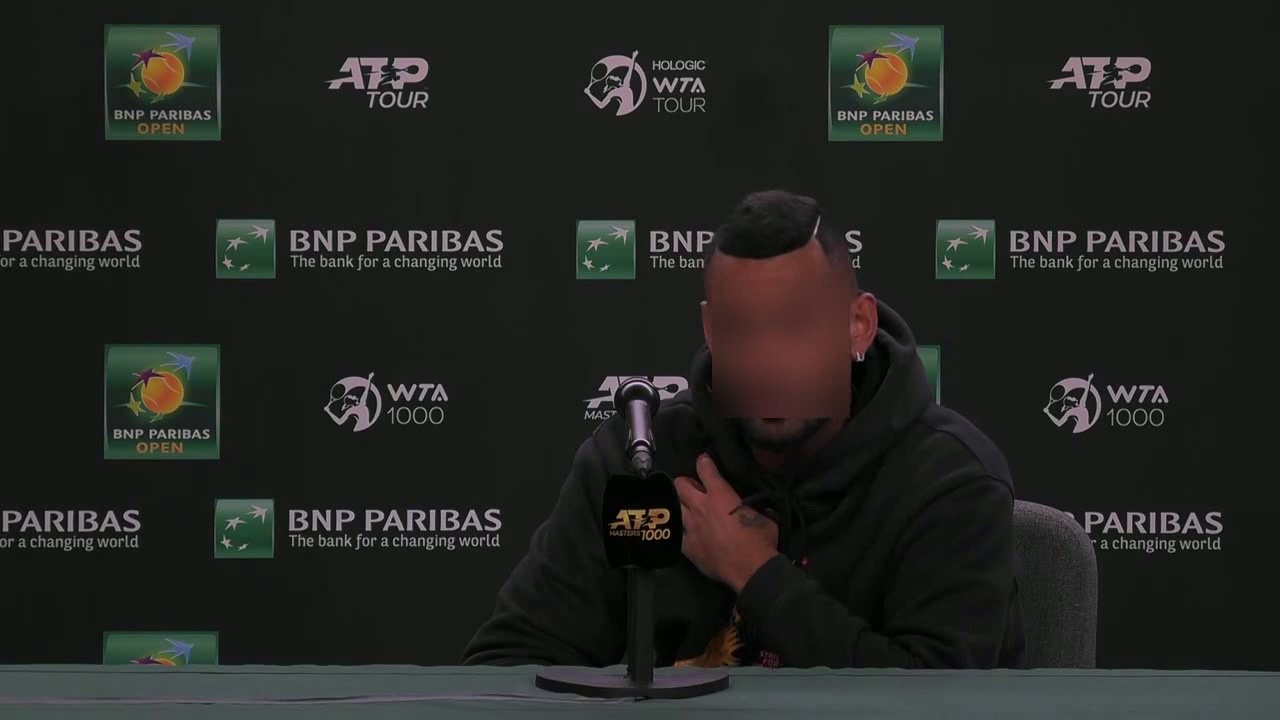}&\includegraphics[width=0.33\linewidth]{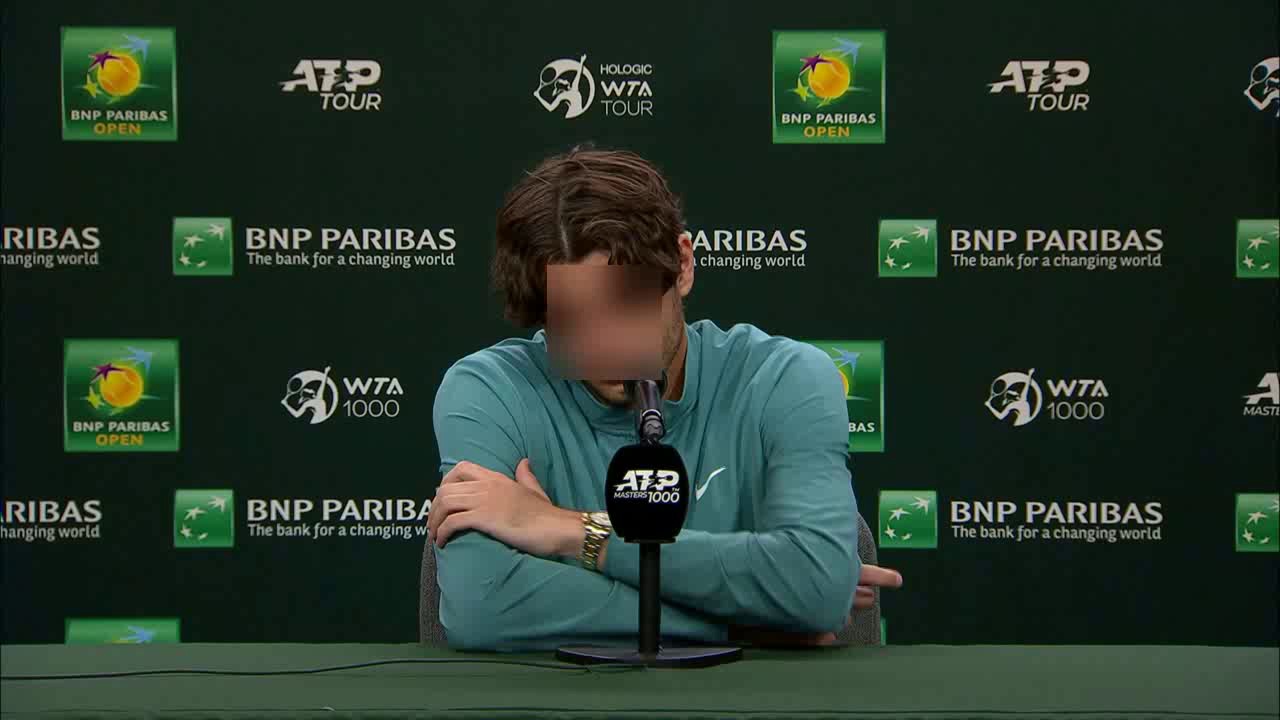}\\
        N12 Playing or adjusting hair&N14 Touching suprasternal notch&N16 Folding arms\\
        \includegraphics[width=0.33\linewidth]{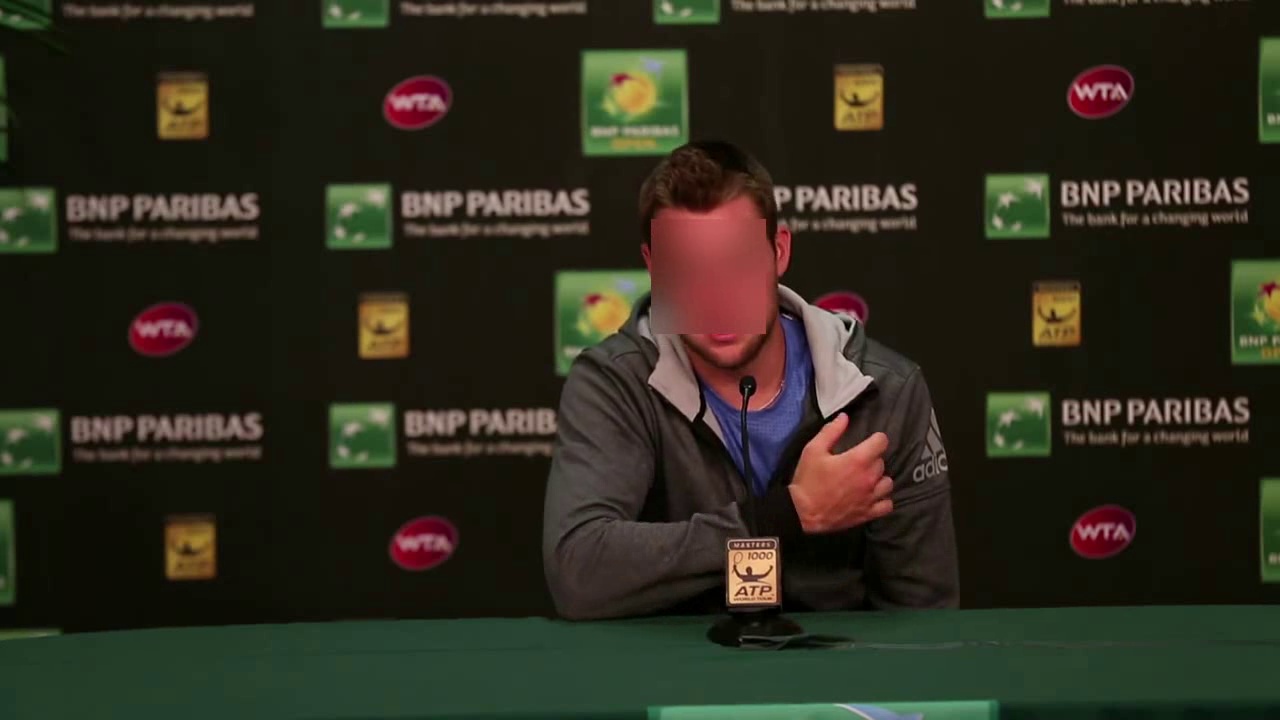}&\includegraphics[width=0.33\linewidth]{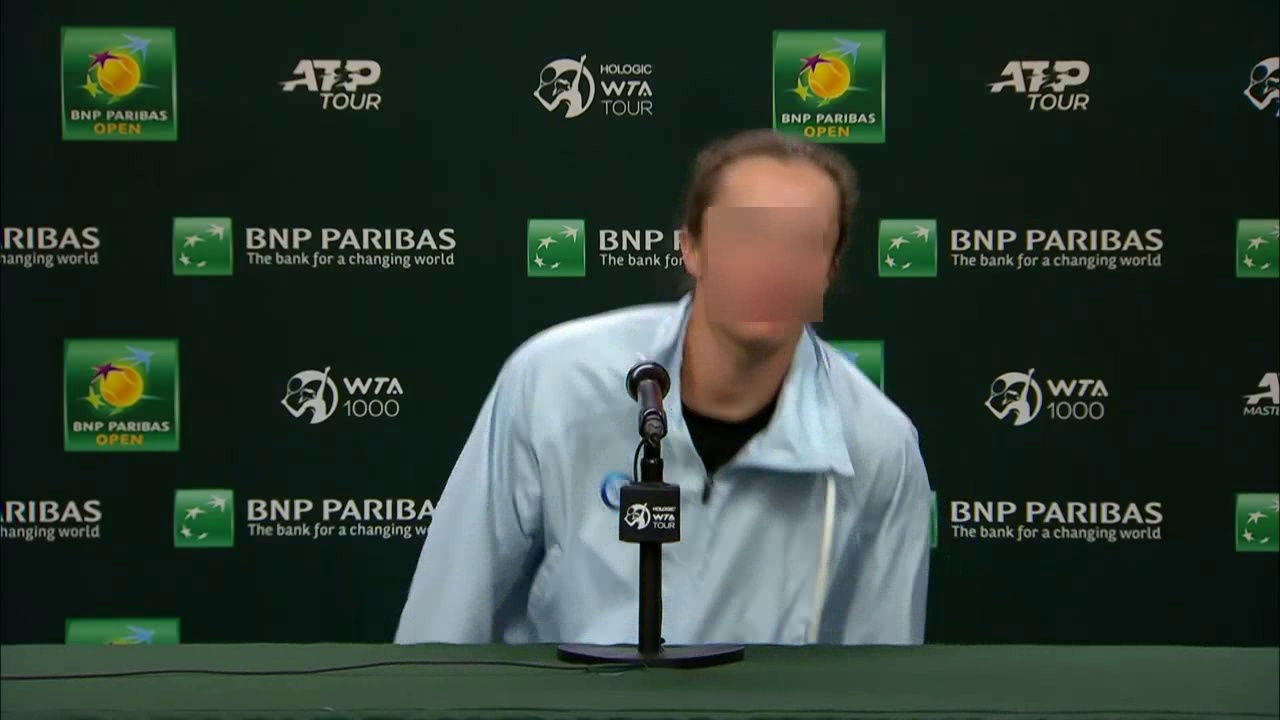}&\includegraphics[width=0.33\linewidth]{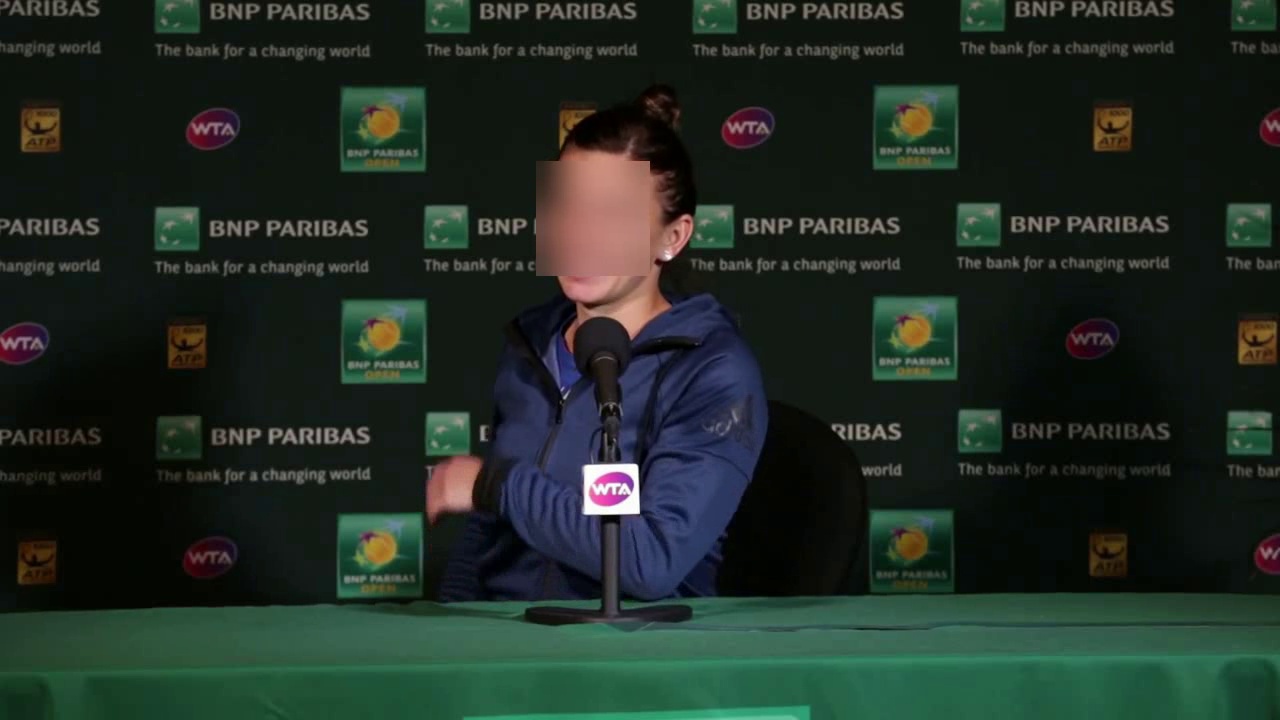}\\
        N17 Dustoff clothes&N19 Moving torso&N20 Sit straightly\\
        \includegraphics[width=0.33\linewidth]{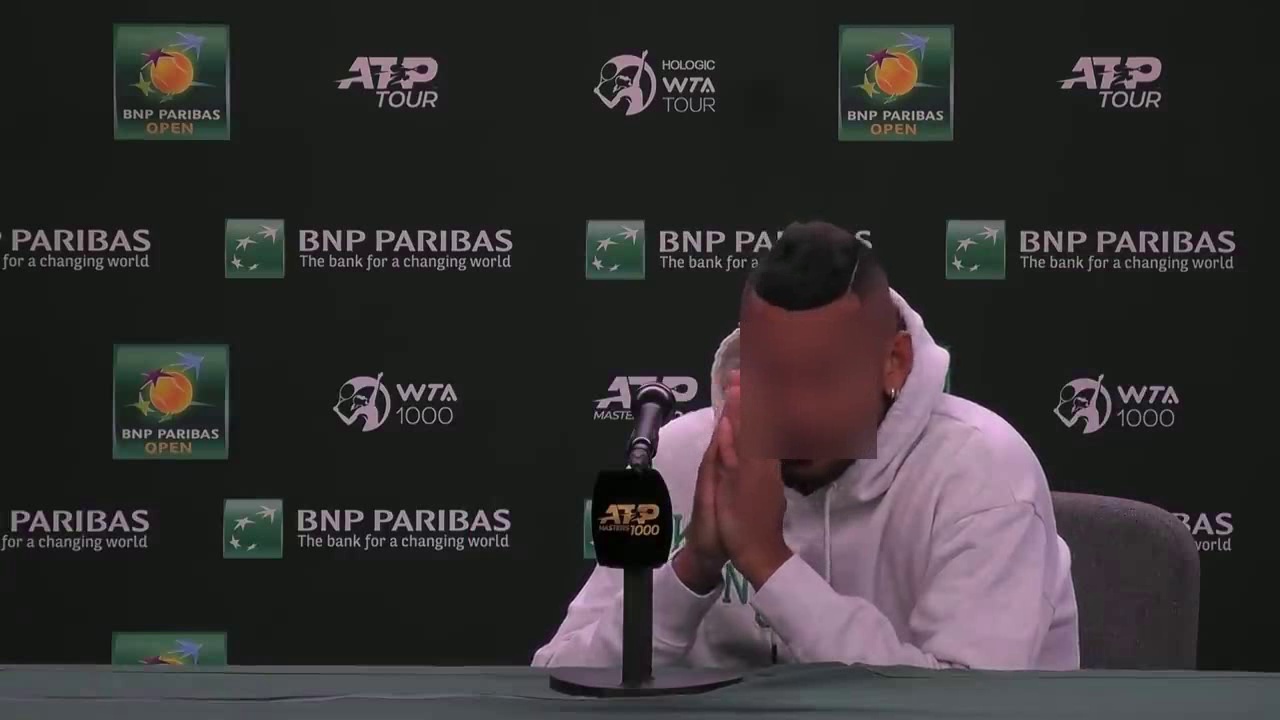}&\includegraphics[width=0.33\linewidth]{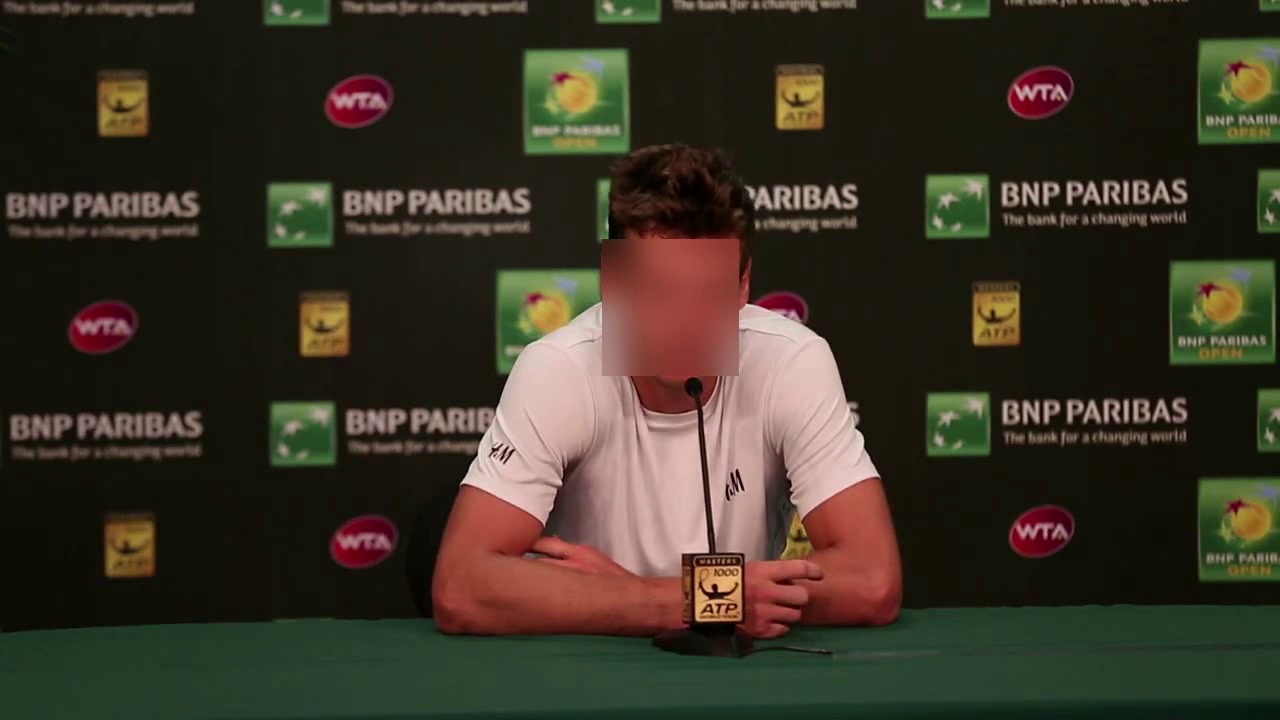}&\includegraphics[width=0.33\linewidth]{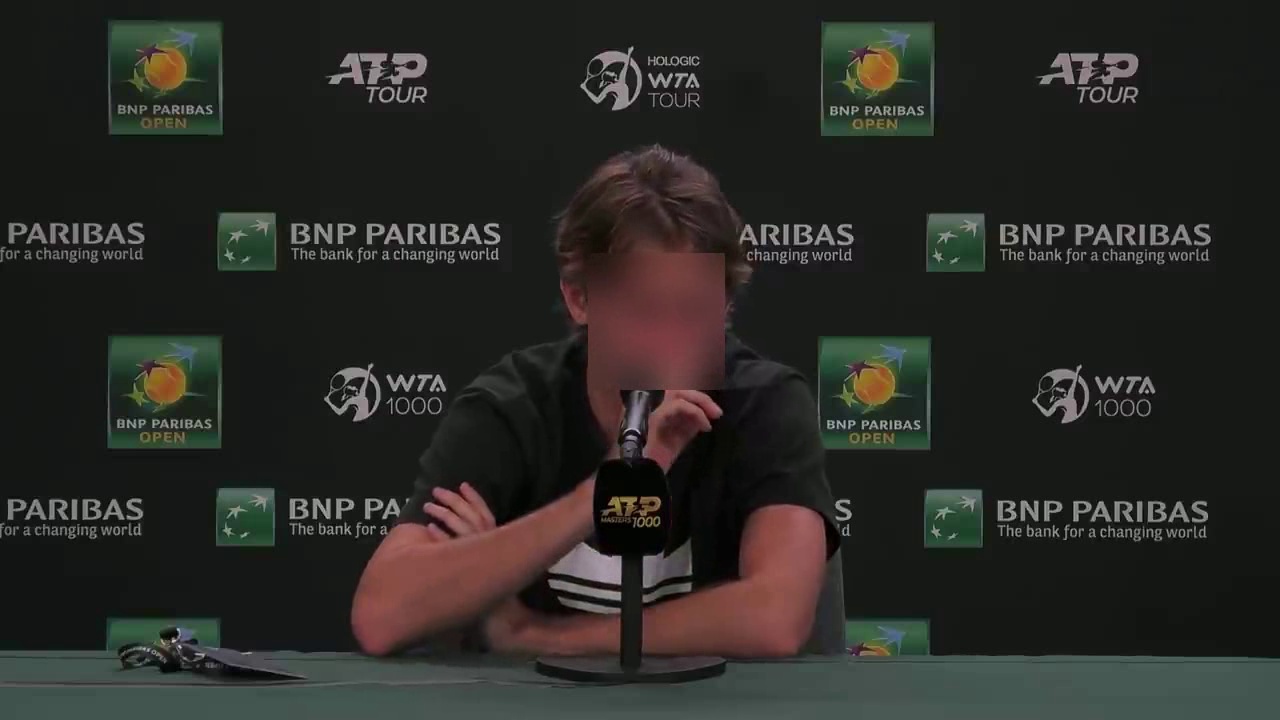}\\
        N24 Minaret gesture&N30 Shake double shoulders&N34 Touching nose\\
    \end{tabular}
    \caption{Selected samples of non-facial body language with the masked face of the proposed dataset \textit{EALD}.}
    \label{fig:samples}
\end{figure}

To solve the aforementioned limitations of the existing datasets, we construct and propose a dataset for Emotion Analysis in Long-sequential and De-identity videos called \textit{EALD}. More precisely, we collect videos containing "post-match press" scenarios in which a professional athlete is given several rounds of Q\&A with reporters after a tough match. As a result of the match, winning or losing is a natural emotion trigger, leading to positive or negative emotional states of the player being interviewed. In addition, we annotated the Non-Facial Body Language (NFBL) of athletes during the interview. Because NFBL has been proven that it can be a crucial clue for understanding human hidden emotions in psychological studies~\cite{loi2013recognition,abramson2021social,aviezer2012body}. Then, We use McAdams~\cite{patino2020speaker} and face detection~\footnote{http://dlib.net/python/} for speech and facial de-identification. Finally, we evaluate various single models (e.g, video and audio) and the Multimodal Large Language Model (MLLM) (e.g., Video-LLaMA~\cite{damonlpsg2023videollama}) on the proposed dataset \textit{EALD}. The experimental results demonstrate that MLLM surpasses other supervised single-modal models, even in the zero-shot scenario. In summary, the contributions of this paper can be concluded as follows:

\begin{itemize}
  \item We construct and propose \textit{EALD} dataset. It bridges the gap in existing datasets in the aspect of de-identity long sequential video emotion analysis.
  \item We provided a benchmark evaluation of the proposed dataset for future research.
  \item We validate that NFBL is an effective and identity-free clue for emotion analysis.
\end{itemize}

\section{EALD dataset}
In this section, we will present the construction and detailed information of the proposed dataset \textit{EALD}.

\subsection{Motivation}
As shown in Table 1, there is a gap between existing datasets and real-life applications: 1) sensitive biological signals involved in their datasets, such as facial and speech. It is hard for the existing datasets to meet the needs for privacy protection in real life; 2) pay more attention to short sequential video emotion analysis but overlook long sequential video. In fact, long sequential video is crucial to emotion analysis. Next, we will show how to construct the proposed dataset \textit{EALD}. And later, we will demonstrate how \textit{EALD} can effectively bridge this gap.
\subsection{Dataset Construction}\label{sec:Dataset Construction}

\subsubsection*{\textbf{Data collection}}
The videos of athletes participating in post-game interviews are a suitable data source. First, the outcome of the game, whether it's a victory or a loss, serves as a natural catalyst for emotions, eliciting either positive or negative states in the interviewed player. Second, the players had no (or very little) time to prepare because the press conference would be held immediately after the game, and he or she needed to respond to the questions rapidly. Unlike acting in movies or series, athletes' NFBL is natural. Third, the duration of the post-game interview videos is commonly relatively long, which meets the need of our target, namely, long sequential emotion analysis. Therefore, we collect videos of post-game interviews in a total of 275 from the Australian Open\footnote{https://www.youtube.com/@australianopen} and BNP Paribas Open\footnote{https://www.youtube.com/@bnpparibasopen} from YouTube.

\subsubsection*{\textbf{Data de-identification}}
As aforementioned, one of the aims of the proposed dataset is to provide identify-free data for emotion analysis. As one may see, the majority of identity information exists in facial and speech signals. Therefore, we perform video and audio de-identification to remove the identity information. Specifically, for video de-identification, we utilize pre-trained face detection Convolutional Neural Network (CNN) model $m_{face\_detection}$ from Dilb\footnote{http://dlib.net/} to implement facial masking for de-identification. Formally, give a video $\mathbf{V} = \{v_1,v_2,...,v_{N_{sample}}\}$ where $v_i$ denote the each frame of video $\mathbf{V}$. Then, we can get the coordinates $C$ by $C = m_{face\_detection}(v_i)$. Finally, we mask the face with Gaussian blur according to the coordinates $C$ of the face. The examples are presented in Fig.~\ref{fig:samples}.

\begin{figure}[htbp]
    \centering
    \small
    \setlength\tabcolsep{0.1pt}
    \begin{tabular}{cc}
        \includegraphics[width=0.48\linewidth]{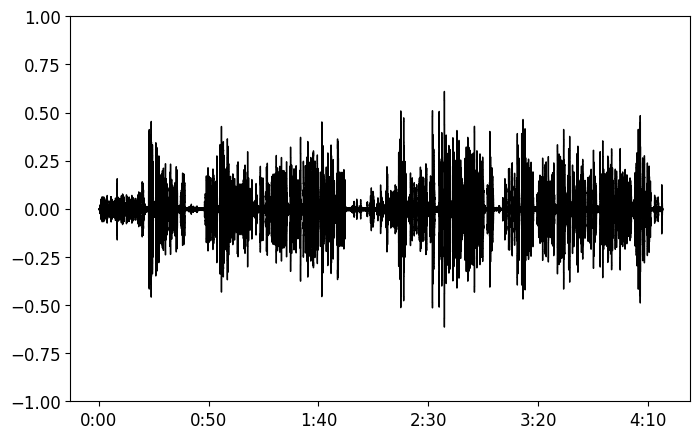}&\includegraphics[width=0.48\linewidth]{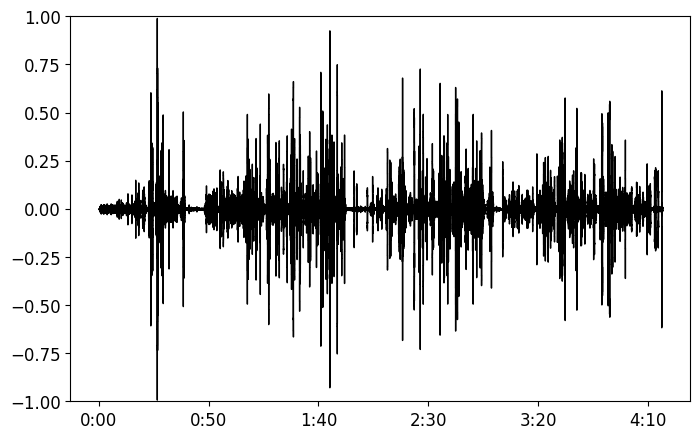}  \\
        (a) waveform of original audio& (b) waveform of de-identity audio\\
        \includegraphics[width=0.48\linewidth]{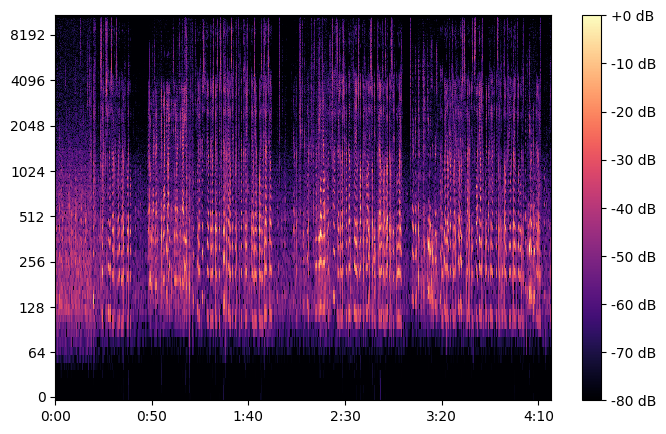}&\includegraphics[width=0.48\linewidth]{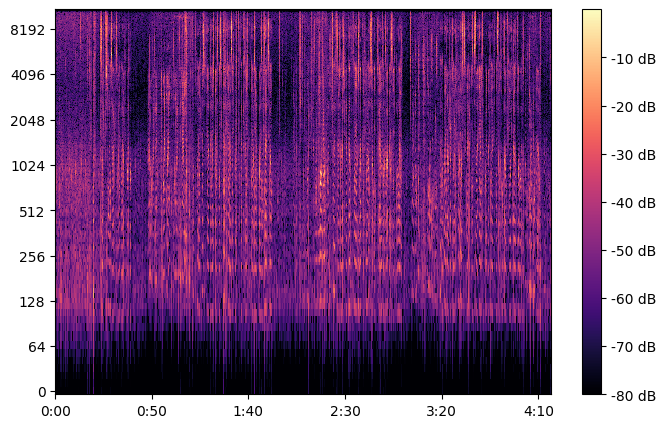}  \\
        (c) spectrogram of original audio& (d) spectrogram of de-identity audio\\
    \end{tabular}
    \caption{Comparsion of the waveform of original with de-identity audio of a sample (sample id 275) of \textit{EALD}}
    \label{fig:audio}
\end{figure}

Regarding audio de-identification, our goal is to remove the identity information in the audio while preserving the emotions. Therefore, we follow the method described in \cite{tomashenko2024voiceprivacy} and utilize the McAdams coefficient-based approach \cite{mcadams1984spectral} to ensure broad applicability and ease of implementation. Specifically, the process of anonymization using the McAdams coefficient is as follows. For an input audio signal $\mathbf{A} = \{a_1,a_2,...,a_{N_{sample}}\}$ with a sampling frequency $f_s$, we divide it into overlapping frames of length \(T_{win}\) milliseconds and step size \(T_{shift}\) milliseconds, denoted as \(\hat{A} = \{\hat{a}_1, \hat{a}_2, ..., \hat{a}_{N_{frame}}\}\). These frames are then processed with a Hanning window, resulting in windowed frame signals \(\hat{S}^w =\{\hat{a}^w_1, \hat{a}^w_2, ..., \hat{a}^w_{N_{frame}}\}\). Then, we utilize Linear Predictive Coding (LPC) analysis and Transfer Function to Zeroes, Poles, and Gain (tf2zpk) to obtain the corresponding poles \(P = \{p_1, p_2, ..., p_{N_{frame}}\}\).
\begin{equation}
    c_i, r_i = \operatorname{LPC}(\hat{s}^w_i, lp\_order),
\end{equation}
\begin{equation}
    z_i, p_i, k_i = \operatorname{tf2zpk}(c_i),
\end{equation}
where \(lp\_order\) is the order of the LPC analysis, $c_i, r_i$ represent the LPC
coefficients and the residuals. Then, we adjust the angle \(\theta_i\) of the poles $p_i$ using the McAdams coefficient \(\lambda\) to obtain the new angle \(\theta_i^{new} = \theta_i^{\lambda}\). Then, based on the new angle \(\theta_i^{new}\) and the original amplitude of the poles, we calculate the new poles \(p_i^{new}\). Then, following the reverse process, \(A^{new}\) is reconstructed. The waveform example of de-identity audio is presented in Fig.~\ref{fig:audio}. 

\subsubsection*{\textbf{Data annotation}}
After data collection and de-identification, each video sequence needs to be annotated with NFBL. According to a study on human behavior~\cite{navarro2008every}, NFBL includes self-manipulations (e.g., scratching the nose and touching the ear), manipulation of objects (e.g., playing with rings, pens, and papers), and self-protection behaviors (e.g., rubbing the eyes, folding arms, or rhythmically moving legs). The defined NFBL classes are presented in Fig.~\ref{fig:Histogram}. The annotation job is carried out in the following stages: First, the NFBL is identified from the collected videos, including start and end times, as well as the NFBL type (clip-level) by trained annotators. We consider "positive" and "negative" as two emotional categories to start with, and the labels are based on the objective fact that the game is won (positive emotion) or lost (negative emotion). After the initial annotation, we perform a review to ensure the accuracy of the labels.

\begin{table}[htbp]
    \centering
    \caption{Properties of the proposed \textit{EALD}.}
    \begin{tabular}{lr}\hline\hline
         Properties&\textit{EALD}\\\hline\hline
         Number of videos&275\\
         Number of annotated NFBL&16180\\
         Resolution&1280 $\times$ 720\\
         Frame rate&32\\
         Subjects&40 Male, 42 Female\\
         Nationality&30\\
         Total duration&32.15 hours\\
         Average duration&7.02 mins\\\hline\hline
    \end{tabular}
    \label{tab:Properties}
\end{table}
\subsection{Dataset Statistics and Properties}

\begin{figure*}
    \centering
    \includegraphics[width=0.6\linewidth]{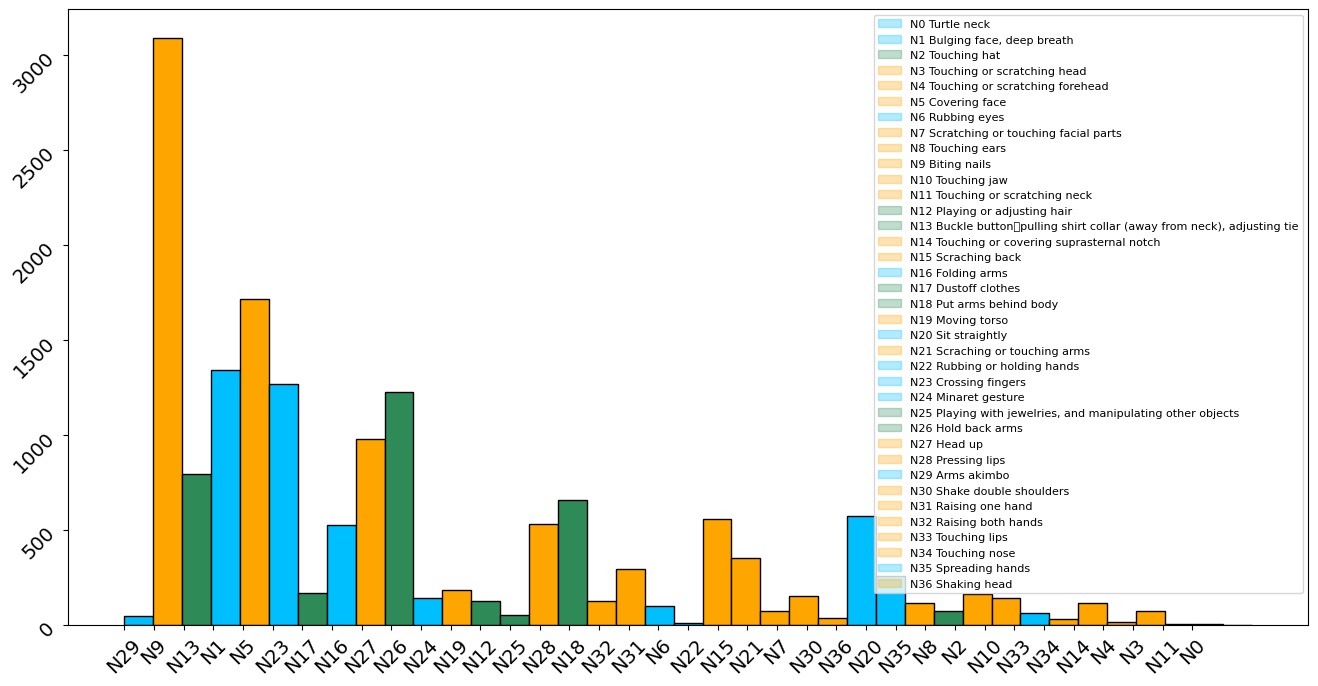}
    \caption{The distribution of various non-facial body language in \textit{EALD}. The y-axis represents the frequency, and the x-axis represents the NFBL. The orange, green, and blue colors represent different types of NFBL, namely, Self-manipulations, Manipulation (touching) objects, and Self-protection behaviors, respectively. Viewed digitally and zoom-in may be better.}
    \label{fig:Histogram}
\end{figure*}

In the proposed dataset \textit{EALD}, we collect 275 (74 Lost, 201 Won) post-game interview videos as shown in Table.~\ref{tab:Properties}. The difference between the proposed dataset and existing datasets can be summarized as follows: 1) long sequential data. To meet our needs for long sequential emotion analysis, the average duration of collected videos is 7.02 minutes. Each video has a resolution of 1280 $\times$ 720 of 32 Frame Per Second (FPS); 2) identity-free: We remove the identity information for the video and audio as described in Sec.~\ref{sec:Dataset Construction}; 3) diversity: The expression for different people may be different because of cultural background, particularly in body language. It is worth emphasizing that the proposed \textit{EALD} is well diverse in the aspects of gender and nationality. For example, the athletes in the collected video from different countries around the world (e.g., Australia, China, Japan, Slovak, South Africa, USA). More importantly, the gender of the athletes is well-balanced (40 Male, 42 Female).

Regarding the annotation of NFBL, we provide 16,180 clips of various NFBL with timestamps in the proposed dataset. The clips range from 0.08 seconds to 184.50 seconds. The distribution of different NFBL is shown in Fig.~\ref{fig:Histogram}. As one may observe, the proportion of N9 (Biting nails) and N5 (Covering face) is significantly higher than that of other categories.

\section{EALD-MLLM}
\begin{figure*}[h!]
    \centering
    \includegraphics[width=0.9\linewidth]{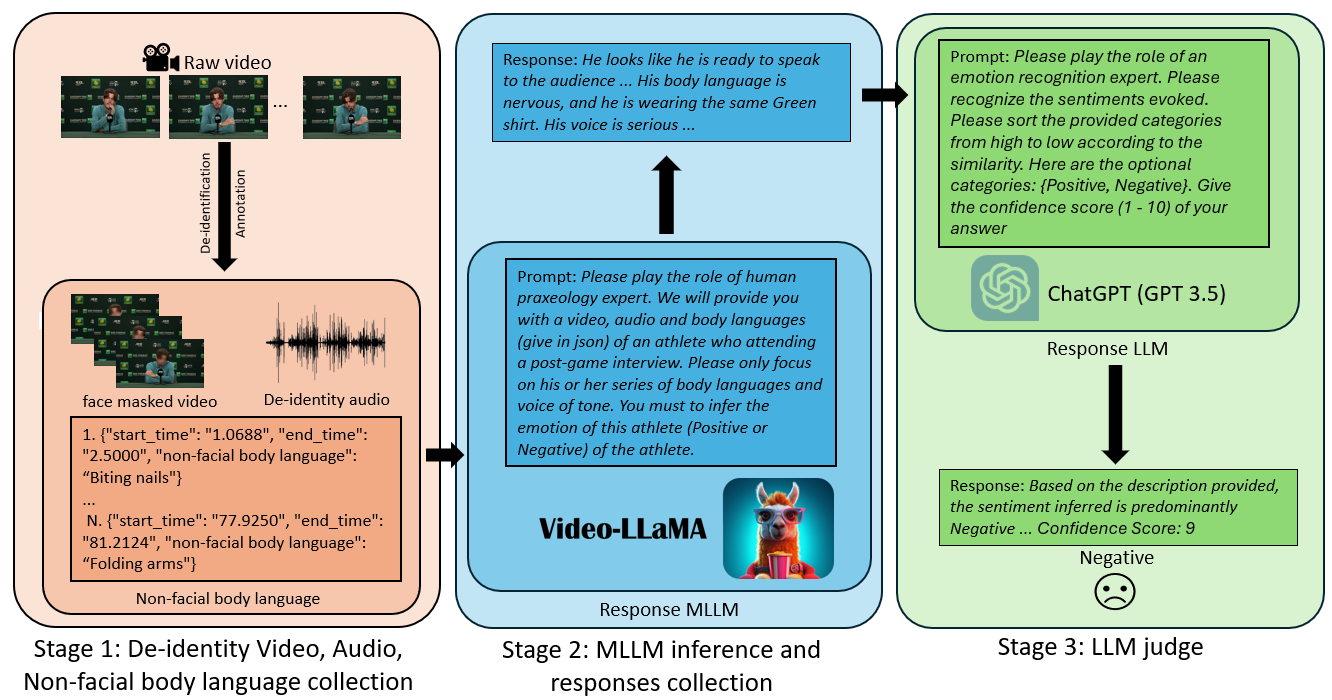}
    \caption{The pipeline of the proposed EALD-MLLM.}
    \label{fig:Framework}
\end{figure*}

\begin{table*}[htbp]
    \centering
    \setlength\tabcolsep{2pt}
    \caption{Ablation study of the \texttt{EALD-MLLM} on full \textit{EALD}.}
    \begin{tabular}{lll|rrrr}\hline\hline
        Video&Audio&NFBL&Accuracy(\%)$\uparrow$&F-score(\%)$\uparrow$&Precision(\%)$\uparrow$&Confidence score$\uparrow$\\\hline
        \checkmark&&&49.09&63.91&66.31&6.08\\
        \checkmark&\checkmark&&53.45&67.51&68.91&6.83\\
        \checkmark&\checkmark&\checkmark&\textbf{58.54}&\textbf{70.77}&\textbf{68.65}&\textbf{7.29}\\
        \hline\hline
    \end{tabular}
    \label{tab:albation}
\end{table*}
\begin{table*}[htbp]
    \centering
    \setlength\tabcolsep{2pt}
    \caption{Comparative experiment on part of \textit{EALD}.}
    \begin{tabular}{l|rrr|rrr}\hline\hline
        Method&Pretraining dataset&Modality&Zero-shot?&Accuracy(\%)$\uparrow$&F-score(\%)$\uparrow$&Precision(\%)$\uparrow$\\\hline
        SlowFast~\cite{feichtenhofer2019slowfast}&K400&Video&$\times$&52.70&42.62&54.10\\
        TSM~\cite{lin2019tsm}&ImageNet+K400&Video&$\times$&51.35&60.87&50.91\\
        TimeSformer~\cite{bertasius2021space}&ImageNet+K400&Video&$\times$&48.65&61.22&49.18\\
        Video-Swin~\cite{liu2022video}&ImageNet+K400&Video&$\times$&51.35&63.27&50.82\\
        BEATs~\cite{chen2022beats}&AudioSet&Audio&$\times$& 50.00 & \textbf{66.67} & 50.00\\
         EALD-MLLM (ours)&Webvid+VideoChat&Video + Audio + NFBL&\checkmark&\textbf{58.10}&\textbf{66.67}&\textbf{55.35}\\\hline\hline
    \end{tabular}
    \label{tab:comparative}
\end{table*}

For humans, it is possible to identify the emotions of others through the visual system, the auditory system, or by combining both. Recently, Multimodal Large Language Models (MLLM) have shown their robust performance in many sub-stream tasks. MLLM combines the capabilities of large language models with the ability to understand content across multiple modalities, such as text, images, audio, and video. Next, we will introduce an MLLM-based solution for emotion analysis.

\subsection{Framework}\label{sec:Framework}
The framework of the proposed EALD-MLLM solution for emotion analysis is illustrated in Fig.\ref{fig:Framework}. As we can see, the proposed solution consists of three stages. Formally, given a video $\mathbf{V}$, corresponding audio $\mathbf{A}$ and corresponding NFBL $\mathbf{N}$. In stage one, we perform de-identification to remove the identity information. Thus, the de-identity video $\mathbf{\hat{V}}$ and audio $\mathbf{\hat{A}}$ can be obtained. More details of de-identification can be found in Sec.\ref{sec:Dataset Construction}.

\subsubsection*{\textbf{MLLM inference}}
In stage two, we employ Video-LLaMA~\cite{damonlpsg2023videollama} as our MLLM for emotion analysis inference, utilizing de-identity video $\mathbf{\hat{V}}$, audio $\mathbf{\hat{A}}$, and NFBL $N$ as inputs. We chose Video-LLaMA because it aligns different modality data (e.g., video, audio, and text), meeting our requirements. Specifically, Video-LLaMA incorporates a pre-trained image encoder into a video encoder known as the Video Q-former, enabling it to learn visual queries, and utilizes ImageBind~\cite{girdhar2023imagebind} as the pre-trained audio encoder, introducing an Audio Q-former to learn auditory queries. Finally, it feeds visual and auditory queries to the pre-trained large language model LLaMA~\cite{touvron2023llama} to generate responses.

Since $\mathbf{MLLM}$ is determined, we simply input de-identity video $\mathbf{\hat{V}}$ and audio $\mathbf{\hat{A}}$ into $\mathbf{MLLM}$. Specifically, we uniformly sample $M$ frames from $\mathbf{\hat{V}}$ to form visual input $\mathbf{\bar{V}} = \{ \hat{v}_1, \hat{v}_2, ..., \hat{v}_m \}$, where $\hat{v}_i$ denotes a frame in $\mathbf{\hat{V}}$. For audio, we sample $S$ segments of 2-second audio clips. Next, we convert each of these 2-second audio clips into spectrograms $\bar{a}_i$ using 128 mel-spectrogram bins to form audio input $\mathbf{\bar{A}} = \{ \bar{a}_1, \bar{a}_2, ..., \bar{a}_s \}$. Regarding NFBL $N$, we utilize it as text input. Finally, the response $R$ is $\mathbf{MLLM}(\mathbf{\bar{V}},\mathbf{\bar{A}}, N)$. It is noted that proposing another new MLLM model is not our goal; instead, we aim to utilize MLLM for long-sequential emotion analysis. Therefore, we choose to use the existing model Video-LLaMA instead of proposing a new MLLM model. Additionally, although the current selection of MLLM may not be optimal, it still achieved satisfactory results, as shown in Table.~\ref{tab:comparative}. Thus, we stop further optimizing it. If one may interested in MLLM, please refer to~\cite{zhang2024vision}.

\subsubsection*{\textbf{LLM judgement}}
Although we can get a reasonable response $R = \mathbf{MLLM}(\mathbf{\bar{V}},\mathbf{\bar{A}}, N)$, $\mathbf{MLLM}$'s response is more about descriptive content than emotion analysis. We think the response is related to the instruction dataset. Because Video-LLaMA is instruction-tuned on Visual Question Answering (VQA) datasets while not on emotion analysis datasets. However, the generated response $R$ is still helpful for emotion analysis because it contains descriptive information about visual, audio, and body language. Therefore, we opt for ChatGPT~\cite{ouyang2022training} from emotion estimation based on the given response $R$. More precisely, we utilize gpt-3.5-turbo-0125 to generate a response that includes the estimation of emotion and the confidence score. The used prompt for ChatGPT is present in Fig.\ref{fig:Framework} (Stage 3).

\begin{figure*}[!h]
    \centering
    \includegraphics[width=0.6\linewidth]{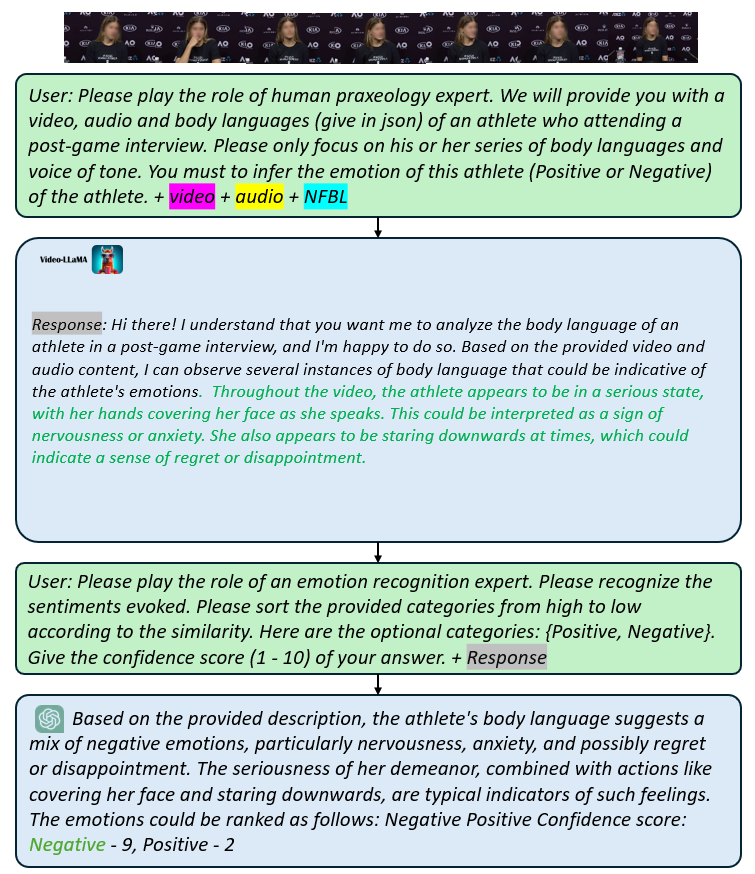}
    \caption{Example case of with Non-facial Body Language}
    \label{fig:Usercase1}
\end{figure*}

\begin{figure*}[!h]
    \centering
    \includegraphics[width=0.6\linewidth]{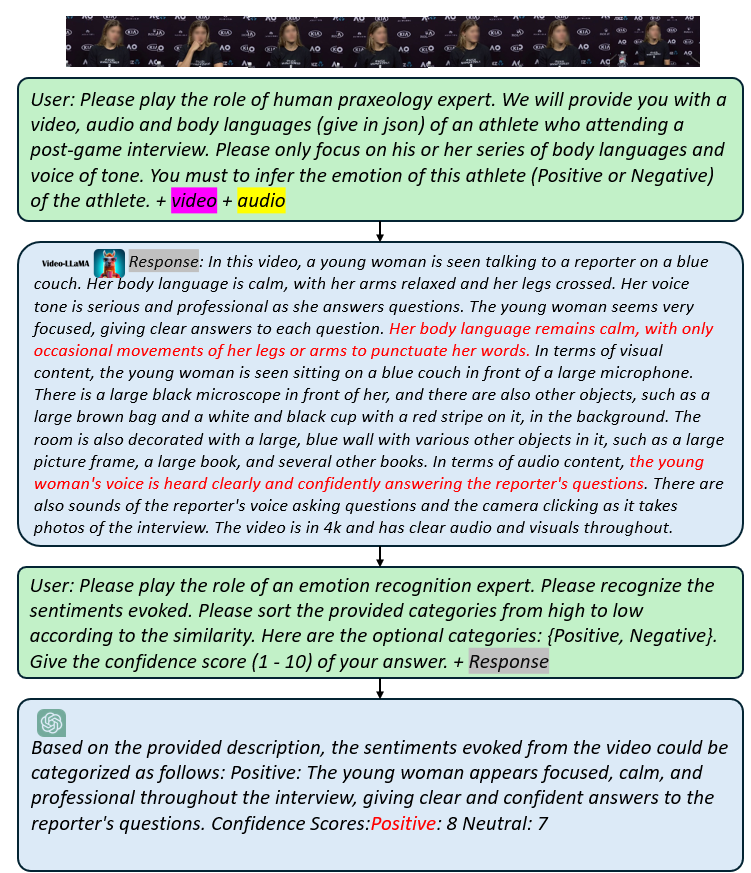}
    \caption{Example case of without Non-facial Body Language. }
    \label{fig:Usercase2}
\end{figure*}
\section{Experimental Result}

In this section, we will introduce the ablation study and comparative study of the proposed MLLM solution \texttt{EALD-MLLM}.

\subsection{Experimental settings}

\subsubsection*{\textbf{Metrics}}
As the emotion analysis \textit{EALD} can be regarded as a binary classification problem, we employ accuracy, precision, and F-score as metrics to assess the performance of different models, as follows:
\begin{equation}
    \text{Accuracy} = \frac{TP + TN}{TP + TN + FP + FN},
\end{equation}
\begin{equation}
    \text{Recall} = \frac{TP}{TP + FN},
\end{equation}
\begin{equation}
    \text{Precision} = \frac{TP}{TP + FP},
\end{equation}
\begin{equation}
    \text{F1-score} = \frac{2 \times \text{Precision} \times  \text{Recall}}{\text{Precision} + \text{Recall}},
\end{equation}
where True Positives (TP), True Negatives (TN), False Positives (FP), and False Negatives (FN) represent the instances accurately predicted as positive, instances correctly predicted as negative, indicate instances inaccurately predicted as positive, and instances inaccurately predicted as negative respectively.

\subsubsection*{\textbf{Inference}} 

As described in Sec.\ref{sec:Framework}, the proposed \texttt{EALD-MLLM} solution for emotion analysis utilizes Video-LLaMA~\cite{damonlpsg2023videollama} and ChatGPT~\cite{ouyang2022training} without training. For the inputs of Video-LLaMA, we uniformly sample 32 frames for each video and sample audio clips every 2 seconds for each audio. As for NFBL, we directly use the annotation of NFBL and input it as the text form for Video-LLaMA.

\subsection{Ablation Study}
Before comparing the \texttt{EALD-MLLM} with other approaches, three key questions need to be addressed: 1) Do we need multimodal data for long-sequential video emotion analysis? 2) Is NFBL helpful for long-sequential video emotion analysis? 3) Does long sequential emotion analysis more accurately reflect genuine emotions? Thus, we conduct an ablation study to investigate these three questions. Since all experiments in the ablation study are conducted under the zero-shot scenario, we use the full data of the proposed \textit{EALD}.

\subsubsection*{\textbf{Dose multimodal data beneficial for emotion analysis?}}
We begin with video input alone as the base model, then gradually integrate audio and NFBL. As shown in Table~\ref{tab:albation}, relying solely on visual cues for detecting emotions proves challenging, with an accuracy below 50\%. In contrast, incorporating audio data significantly improves the model's accuracy by about 4\%. Therefore, we believe that both video and audio are crucial modalities as they can provide rich information for long-sequential emotion analysis. Utilizing multimodal techniques that combine the features of video and audio is advantageous for emotion analysis.

\subsubsection*{\textbf{Dose non-facial body language Helpful?}}
As shown in Table \ref{tab:albation}, combining NFBL can significantly improve performance (by 5\% in accuracy). Additionally, it enhances the confidence score, which represents the descriptive content generated by Video-LLaMA and provides more informative insights for emotion analysis.

\subsubsection*{\textbf{Does emotion analysis over a long sequential period more accurately reflect people's genuine emotions?}}
As shown in the supplementary materials, when we compare instantaneous (short sequence) and long sequential video emotion analysis, long-term analysis is better able to reflect people's actual emotional states comprehensively. Short-term recognition may be influenced by specific moments, whereas long-term observation can better capture emotional fluctuations and trends, thus providing more accurate emotional analysis. Long-term emotion recognition also allows for a deeper understanding of the context and underlying factors influencing individuals' emotional states. This contextual richness further enhances the accuracy and depth of the emotional analysis over time. Due to limited space, please refer to the details in the supplementary materials.

\subsection{Benchmark Evaluation}
To validate the effectiveness of the proposed \texttt{EALD-MLLM} solution, we selected several single-modal models for comparison: 1) for video-based models, we chose Video-Swin-Transformer (Video-Swin)~\cite{liu2022video}, TSM~\cite{lin2019tsm}, TimeSformer~\cite{bertasius2021space}, and SlowFast~\cite{feichtenhofer2019slowfast}; 2) for audio-based models, we selected BEATs~\cite{chen2022beats}. Typically, these methods require training before they can be tested on the proposed dataset. Thus, we randomly selected data from the dataset for the comparative experiment to reduce the impact of data imbalance. Specifically, we used 72 videos (36 Negative, 36 Positive) for training and 74 videos (37 Negative, 37 Positive) for testing. The video IDs of the selected dataset can be found in the Appendix. Considering the size of the training dataset is relatively small, we opted for linear probing to train these models. Thus, we only trained the last classification layer and froze the rest of the pre-trained layers. Video-Swin~\cite{liu2022video}, TSM~\cite{lin2019tsm}, TimeSformer~\cite{bertasius2021space} and are pre-trained on the Kinetics-400~\cite{carreira2017quo} and ImageNet~\cite{deng2009imagenet}. SlowFast~\cite{feichtenhofer2019slowfast} is pre-trained on the Kinetics-400~\cite{carreira2017quo}.Then, we employ an AdamW optimizer for 30 epochs using a cosine decay learning rate scheduler and 2.5 epochs of linear warm-up. The audio recognition model BEATs~\cite{chen2022beats} is pre-trained on the AudioSet~\cite{gemmeke2017audio}. Then, we employ an AdamW optimizer for 30 epochs with EarlyStopping. It is noted that our audio files typically have a duration of about 10 minutes, which can be unacceptable for conventional audio classification models to handle. To enhance the feasibility of training, we resample the original audio from a sampling rate of 16,000 Hz to 320 Hz.

The results displayed in Table \ref{tab:comparative} indicate that classic video or audio recognition models exhibit relatively poor performance, achieving an accuracy of approximately 50\%. This also proves that using only one modality is not suitable for long-sequential emotion analysis. In contrast, our proposed solution, \texttt{EALD-MLLM}, achieves superior performance, boasting a 5\% increase in accuracy. The qualitative results are present in Fig.~\ref{fig:Usercase1} and Fig.~\ref{fig:Usercase2}.

\section{Limitation}

\begin{figure}
    \centering
    \includegraphics[width=\linewidth]{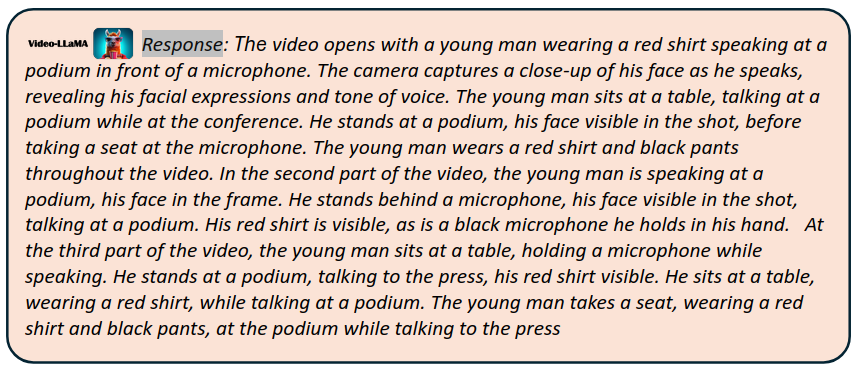}
    \caption{Example case of failure MLLM response. The response lacks the information for emotion analysis. }
    \label{fig:failure}
\end{figure}

As one can see, although the proposed \texttt{EALD-MLLM} outperforms other models, its performance is not very high, as shown in Table~\ref{tab:comparative}. This also proves that the proposed \texttt{EALD} dataset designed for long-sequential emotion analysis in this paper is challenging. In Table~\ref{tab:albation}, we validate that Non-Facial Body Language (NFBL) is an important clue for long-sequential emotion analysis. However, we use the annotation of NFBL directly. To our knowledge, detecting these NFBLs in real applications is not a simple task, as they often occur quickly and are difficult to recognize. Therefore, we plan to study the detection of NFBLs in the future. Furthermore, the \texttt{EALD-MLLM} approach follows a two-stage methodology. Because the employed MlLM (Video-LLaMA) is specifically designed for general purposes and is not optimized for emotion analysis. As shown in Fig.\ref{fig:failure}, in some cases, the generated response by MLLM does not provide useful information for emotion analysis while paying more attention to the environment information. Therefore, we have to use another LLM (ChatGPT) to refine the response and estimate the final emotion. We plan to propose an end-to-end MLLM model for long-sequential emotion analysis in the future.

\section{Conclusion}
In this paper, we introduce a novel dataset for Emotion Analysis in Long-sequential and De-identified termed \textit{EALD} to solve the limitations of the existing datasets: 1) lack of long sequential video for emotion analysis; 2) sensitive data (e.g., de-identity facial, speech, and ECG signals) involved. The \textit{EALD} dataset comprises 275 videos, each with an average duration of 7 minutes. It is worth emphasizing that we perform de-identification processing on all videos and audio. In addition to providing annotations of the overall emotional state of each video, we also provide the Non-Facial Body Language (NFBL) annotations for each player. Furthermore, we propose a solution utilizing a Multimodal Large Language Model (MLLM) called \texttt{EALD-MLMM}. Experimental results demonstrate: 1) MLLMs can achieve comparable performance to supervised single-modal models, even in zero-shot scenarios; 2) NFBL serves as an important identity-free cue in the analysis of emotions over long-sequential.

\bibliographystyle{ACM-Reference-Format}
\bibliography{main}


\begin{thebibliography}{47}


\ifx \showCODEN    \undefined \def \showCODEN     #1{\unskip}     \fi
\ifx \showDOI      \undefined \def \showDOI       #1{#1}\fi
\ifx \showISBNx    \undefined \def \showISBNx     #1{\unskip}     \fi
\ifx \showISBNxiii \undefined \def \showISBNxiii  #1{\unskip}     \fi
\ifx \showISSN     \undefined \def \showISSN      #1{\unskip}     \fi
\ifx \showLCCN     \undefined \def \showLCCN      #1{\unskip}     \fi
\ifx \shownote     \undefined \def \shownote      #1{#1}          \fi
\ifx \showarticletitle \undefined \def \showarticletitle #1{#1}   \fi
\ifx \showURL      \undefined \def \showURL       {\relax}        \fi
\providecommand\bibfield[2]{#2}
\providecommand\bibinfo[2]{#2}
\providecommand\natexlab[1]{#1}
\providecommand\showeprint[2][]{arXiv:#2}

\bibitem[Abramson et~al\mbox{.}(2021)]%
        {abramson2021social}
\bibfield{author}{\bibinfo{person}{Lior Abramson}, \bibinfo{person}{Rotem Petranker}, \bibinfo{person}{Inbal Marom}, {and} \bibinfo{person}{Hillel Aviezer}.} \bibinfo{year}{2021}\natexlab{}.
\newblock \showarticletitle{Social interaction context shapes emotion recognition through body language, not facial expressions.}
\newblock \bibinfo{journal}{\emph{Emotion}} \bibinfo{volume}{21}, \bibinfo{number}{3} (\bibinfo{year}{2021}), \bibinfo{pages}{557}.
\newblock


\bibitem[Aviezer et~al\mbox{.}(2012)]%
        {aviezer2012body}
\bibfield{author}{\bibinfo{person}{Hillel Aviezer}, \bibinfo{person}{Yaacov Trope}, {and} \bibinfo{person}{Alexander Todorov}.} \bibinfo{year}{2012}\natexlab{}.
\newblock \showarticletitle{Body cues, not facial expressions, discriminate between intense positive and negative emotions}.
\newblock \bibinfo{journal}{\emph{Science}} \bibinfo{volume}{338}, \bibinfo{number}{6111} (\bibinfo{year}{2012}), \bibinfo{pages}{1225--1229}.
\newblock


\bibitem[Baveye et~al\mbox{.}(2015)]%
        {baveye2015liris}
\bibfield{author}{\bibinfo{person}{Yoann Baveye}, \bibinfo{person}{Emmanuel Dellandrea}, \bibinfo{person}{Christel Chamaret}, {and} \bibinfo{person}{Liming Chen}.} \bibinfo{year}{2015}\natexlab{}.
\newblock \showarticletitle{LIRIS-ACCEDE: A video database for affective content analysis}.
\newblock \bibinfo{journal}{\emph{IEEE Transactions on Affective Computing}} \bibinfo{volume}{6}, \bibinfo{number}{1} (\bibinfo{year}{2015}), \bibinfo{pages}{43--55}.
\newblock


\bibitem[Bertasius et~al\mbox{.}(2021)]%
        {bertasius2021space}
\bibfield{author}{\bibinfo{person}{Gedas Bertasius}, \bibinfo{person}{Heng Wang}, {and} \bibinfo{person}{Lorenzo Torresani}.} \bibinfo{year}{2021}\natexlab{}.
\newblock \showarticletitle{Is space-time attention all you need for video understanding?}. In \bibinfo{booktitle}{\emph{ICML}}, Vol.~\bibinfo{volume}{2}. \bibinfo{pages}{4}.
\newblock


\bibitem[Busso et~al\mbox{.}(2008)]%
        {busso2008iemocap}
\bibfield{author}{\bibinfo{person}{Carlos Busso}, \bibinfo{person}{Murtaza Bulut}, \bibinfo{person}{Chi-Chun Lee}, \bibinfo{person}{Abe Kazemzadeh}, \bibinfo{person}{Emily Mower}, \bibinfo{person}{Samuel Kim}, \bibinfo{person}{Jeannette~N Chang}, \bibinfo{person}{Sungbok Lee}, {and} \bibinfo{person}{Shrikanth~S Narayanan}.} \bibinfo{year}{2008}\natexlab{}.
\newblock \showarticletitle{IEMOCAP: Interactive emotional dyadic motion capture database}.
\newblock \bibinfo{journal}{\emph{Language Resources and Evaluation}}  \bibinfo{volume}{42} (\bibinfo{year}{2008}), \bibinfo{pages}{335--359}.
\newblock


\bibitem[Carreira and Zisserman(2017)]%
        {carreira2017quo}
\bibfield{author}{\bibinfo{person}{Joao Carreira} {and} \bibinfo{person}{Andrew Zisserman}.} \bibinfo{year}{2017}\natexlab{}.
\newblock \showarticletitle{Quo vadis, action recognition? a new model and the kinetics dataset}. In \bibinfo{booktitle}{\emph{proceedings of the IEEE Conference on Computer Vision and Pattern Recognition}}. \bibinfo{pages}{6299--6308}.
\newblock


\bibitem[Cavallo et~al\mbox{.}(2018)]%
        {cavallo2018emotion}
\bibfield{author}{\bibinfo{person}{Filippo Cavallo}, \bibinfo{person}{Francesco Semeraro}, \bibinfo{person}{Laura Fiorini}, \bibinfo{person}{Gergely Magyar}, \bibinfo{person}{Peter Sin{\v{c}}{\'a}k}, {and} \bibinfo{person}{Paolo Dario}.} \bibinfo{year}{2018}\natexlab{}.
\newblock \showarticletitle{Emotion modelling for social robotics applications: a review}.
\newblock \bibinfo{journal}{\emph{Journal of Bionic Engineering}}  \bibinfo{volume}{15} (\bibinfo{year}{2018}), \bibinfo{pages}{185--203}.
\newblock


\bibitem[Chen et~al\mbox{.}(2022)]%
        {chen2022beats}
\bibfield{author}{\bibinfo{person}{Sanyuan Chen}, \bibinfo{person}{Yu Wu}, \bibinfo{person}{Chengyi Wang}, \bibinfo{person}{Shujie Liu}, \bibinfo{person}{Daniel Tompkins}, \bibinfo{person}{Zhuo Chen}, {and} \bibinfo{person}{Furu Wei}.} \bibinfo{year}{2022}\natexlab{}.
\newblock \showarticletitle{Beats: Audio pre-training with acoustic tokenizers}.
\newblock \bibinfo{journal}{\emph{arXiv preprint arXiv:2212.09058}} (\bibinfo{year}{2022}).
\newblock


\bibitem[Deng et~al\mbox{.}(2009)]%
        {deng2009imagenet}
\bibfield{author}{\bibinfo{person}{Jia Deng}, \bibinfo{person}{Wei Dong}, \bibinfo{person}{Richard Socher}, \bibinfo{person}{Li-Jia Li}, \bibinfo{person}{Kai Li}, {and} \bibinfo{person}{Li Fei-Fei}.} \bibinfo{year}{2009}\natexlab{}.
\newblock \showarticletitle{Imagenet: A large-scale hierarchical image database}. In \bibinfo{booktitle}{\emph{2009 IEEE conference on computer vision and pattern recognition}}. Ieee, \bibinfo{pages}{248--255}.
\newblock


\bibitem[Dhall et~al\mbox{.}(2012)]%
        {dhall2012collecting}
\bibfield{author}{\bibinfo{person}{Abhinav Dhall}, \bibinfo{person}{Roland Goecke}, \bibinfo{person}{Simon Lucey}, {and} \bibinfo{person}{Tom Gedeon}.} \bibinfo{year}{2012}\natexlab{}.
\newblock \showarticletitle{Collecting Large, Richly Annotated Facial-Expression Databases from Movies}.
\newblock \bibinfo{journal}{\emph{IEEE MultiMedia}} \bibinfo{volume}{19}, \bibinfo{number}{3} (\bibinfo{year}{2012}), \bibinfo{pages}{34--41}.
\newblock
\urldef\tempurl%
\url{https://doi.org/10.1109/MMUL.2012.26}
\showDOI{\tempurl}


\bibitem[Dhall et~al\mbox{.}(2015)]%
        {dhall2015video}
\bibfield{author}{\bibinfo{person}{Abhinav Dhall}, \bibinfo{person}{OV Ramana~Murthy}, \bibinfo{person}{Roland Goecke}, \bibinfo{person}{Jyoti Joshi}, {and} \bibinfo{person}{Tom Gedeon}.} \bibinfo{year}{2015}\natexlab{}.
\newblock \showarticletitle{Video and image based emotion recognition challenges in the wild: Emotiw 2015}. In \bibinfo{booktitle}{\emph{ACM on International Conference on Multimodal Interaction}}. \bibinfo{pages}{423--426}.
\newblock


\bibitem[Douglas-Cowie et~al\mbox{.}(2007)]%
        {douglas2007humaine}
\bibfield{author}{\bibinfo{person}{Ellen Douglas-Cowie}, \bibinfo{person}{Roddy Cowie}, \bibinfo{person}{Ian Sneddon}, \bibinfo{person}{Cate Cox}, \bibinfo{person}{Orla Lowry}, \bibinfo{person}{Margaret Mcrorie}, \bibinfo{person}{Jean-Claude Martin}, \bibinfo{person}{Laurence Devillers}, \bibinfo{person}{Sarkis Abrilian}, \bibinfo{person}{Anton Batliner}, {et~al\mbox{.}}} \bibinfo{year}{2007}\natexlab{}.
\newblock \showarticletitle{The HUMAINE database: Addressing the collection and annotation of naturalistic and induced emotional data}. In \bibinfo{booktitle}{\emph{International Conference on Affective Computing and Intelligent Interaction}}. Springer, \bibinfo{pages}{488--500}.
\newblock


\bibitem[El~Ayadi et~al\mbox{.}(2011)]%
        {el2011survey}
\bibfield{author}{\bibinfo{person}{Moataz El~Ayadi}, \bibinfo{person}{Mohamed~S Kamel}, {and} \bibinfo{person}{Fakhri Karray}.} \bibinfo{year}{2011}\natexlab{}.
\newblock \showarticletitle{Survey on speech emotion recognition: Features, classification schemes, and databases}.
\newblock \bibinfo{journal}{\emph{Pattern recognition}} \bibinfo{volume}{44}, \bibinfo{number}{3} (\bibinfo{year}{2011}), \bibinfo{pages}{572--587}.
\newblock


\bibitem[Ezzameli and Mahersia(2023)]%
        {ezzameli2023emotion}
\bibfield{author}{\bibinfo{person}{K Ezzameli} {and} \bibinfo{person}{H Mahersia}.} \bibinfo{year}{2023}\natexlab{}.
\newblock \showarticletitle{Emotion recognition from unimodal to multimodal analysis: A review}.
\newblock \bibinfo{journal}{\emph{Information Fusion}} (\bibinfo{year}{2023}), \bibinfo{pages}{101847}.
\newblock


\bibitem[Fabian Benitez-Quiroz et~al\mbox{.}(2016)]%
        {fabian2016emotionet}
\bibfield{author}{\bibinfo{person}{C Fabian Benitez-Quiroz}, \bibinfo{person}{Ramprakash Srinivasan}, {and} \bibinfo{person}{Aleix~M Martinez}.} \bibinfo{year}{2016}\natexlab{}.
\newblock \showarticletitle{Emotionet: An accurate, real-time algorithm for the automatic annotation of a million facial expressions in the wild}. In \bibinfo{booktitle}{\emph{IEEE Conference on Computer Vision and Pattern Recognition}}. \bibinfo{pages}{5562--5570}.
\newblock


\bibitem[Feichtenhofer et~al\mbox{.}(2019)]%
        {feichtenhofer2019slowfast}
\bibfield{author}{\bibinfo{person}{Christoph Feichtenhofer}, \bibinfo{person}{Haoqi Fan}, \bibinfo{person}{Jitendra Malik}, {and} \bibinfo{person}{Kaiming He}.} \bibinfo{year}{2019}\natexlab{}.
\newblock \showarticletitle{Slowfast networks for video recognition}. In \bibinfo{booktitle}{\emph{Proceedings of the IEEE/CVF international conference on computer vision}}. \bibinfo{pages}{6202--6211}.
\newblock


\bibitem[Fourati and Pelachaud(2014)]%
        {fourati2014emilya}
\bibfield{author}{\bibinfo{person}{Nesrine Fourati} {and} \bibinfo{person}{Catherine Pelachaud}.} \bibinfo{year}{2014}\natexlab{}.
\newblock \showarticletitle{Emilya: Emotional body expression in daily actions database.}. In \bibinfo{booktitle}{\emph{LREC}}. \bibinfo{pages}{3486--3493}.
\newblock


\bibitem[Gemmeke et~al\mbox{.}(2017)]%
        {gemmeke2017audio}
\bibfield{author}{\bibinfo{person}{Jort~F Gemmeke}, \bibinfo{person}{Daniel~PW Ellis}, \bibinfo{person}{Dylan Freedman}, \bibinfo{person}{Aren Jansen}, \bibinfo{person}{Wade Lawrence}, \bibinfo{person}{R~Channing Moore}, \bibinfo{person}{Manoj Plakal}, {and} \bibinfo{person}{Marvin Ritter}.} \bibinfo{year}{2017}\natexlab{}.
\newblock \showarticletitle{Audio set: An ontology and human-labeled dataset for audio events}. In \bibinfo{booktitle}{\emph{2017 IEEE international conference on acoustics, speech and Signal Processing}}. IEEE, \bibinfo{pages}{776--780}.
\newblock


\bibitem[Girdhar et~al\mbox{.}(2023)]%
        {girdhar2023imagebind}
\bibfield{author}{\bibinfo{person}{Rohit Girdhar}, \bibinfo{person}{Alaaeldin El-Nouby}, \bibinfo{person}{Zhuang Liu}, \bibinfo{person}{Mannat Singh}, \bibinfo{person}{Kalyan~Vasudev Alwala}, \bibinfo{person}{Armand Joulin}, {and} \bibinfo{person}{Ishan Misra}.} \bibinfo{year}{2023}\natexlab{}.
\newblock \showarticletitle{Imagebind: One embedding space to bind them all}. In \bibinfo{booktitle}{\emph{Proceedings of the IEEE/CVF Conference on Computer Vision and Pattern Recognition}}. \bibinfo{pages}{15180--15190}.
\newblock


\bibitem[Grimm et~al\mbox{.}(2008)]%
        {grimm2008vera}
\bibfield{author}{\bibinfo{person}{Michael Grimm}, \bibinfo{person}{Kristian Kroschel}, {and} \bibinfo{person}{Shrikanth Narayanan}.} \bibinfo{year}{2008}\natexlab{}.
\newblock \showarticletitle{The Vera am Mittag German audio-visual emotional speech database}. In \bibinfo{booktitle}{\emph{IEEE International Conference on Multimedia and Expo}}. IEEE, \bibinfo{pages}{865--868}.
\newblock


\bibitem[Hakak et~al\mbox{.}(2017)]%
        {hakak2017emotion}
\bibfield{author}{\bibinfo{person}{Nida~Manzoor Hakak}, \bibinfo{person}{Mohsin Mohd}, \bibinfo{person}{Mahira Kirmani}, {and} \bibinfo{person}{Mudasir Mohd}.} \bibinfo{year}{2017}\natexlab{}.
\newblock \showarticletitle{Emotion analysis: A survey}. In \bibinfo{booktitle}{\emph{2017 international conference on computer, communications and electronics (COMPTELIX)}}. IEEE, \bibinfo{pages}{397--402}.
\newblock


\bibitem[Hsu et~al\mbox{.}(2017)]%
        {hsu2017automatic}
\bibfield{author}{\bibinfo{person}{Yu-Liang Hsu}, \bibinfo{person}{Jeen-Shing Wang}, \bibinfo{person}{Wei-Chun Chiang}, {and} \bibinfo{person}{Chien-Han Hung}.} \bibinfo{year}{2017}\natexlab{}.
\newblock \showarticletitle{Automatic ECG-based emotion recognition in music listening}.
\newblock \bibinfo{journal}{\emph{IEEE Transactions on Affective Computing}} \bibinfo{volume}{11}, \bibinfo{number}{1} (\bibinfo{year}{2017}), \bibinfo{pages}{85--99}.
\newblock


\bibitem[Khare et~al\mbox{.}(2023)]%
        {khare2023emotion}
\bibfield{author}{\bibinfo{person}{Smith~K Khare}, \bibinfo{person}{Victoria Blanes-Vidal}, \bibinfo{person}{Esmaeil~S Nadimi}, {and} \bibinfo{person}{U~Rajendra Acharya}.} \bibinfo{year}{2023}\natexlab{}.
\newblock \showarticletitle{Emotion recognition and artificial intelligence: A systematic review (2014--2023) and research recommendations}.
\newblock \bibinfo{journal}{\emph{Information Fusion}} (\bibinfo{year}{2023}), \bibinfo{pages}{102019}.
\newblock


\bibitem[Koelstra et~al\mbox{.}(2011)]%
        {koelstra2011deap}
\bibfield{author}{\bibinfo{person}{Sander Koelstra}, \bibinfo{person}{Christian Muhl}, \bibinfo{person}{Mohammad Soleymani}, \bibinfo{person}{Jong-Seok Lee}, \bibinfo{person}{Ashkan Yazdani}, \bibinfo{person}{Touradj Ebrahimi}, \bibinfo{person}{Thierry Pun}, \bibinfo{person}{Anton Nijholt}, {and} \bibinfo{person}{Ioannis Patras}.} \bibinfo{year}{2011}\natexlab{}.
\newblock \showarticletitle{Deap: A database for emotion analysis; using physiological signals}.
\newblock \bibinfo{journal}{\emph{IEEE Transactions on Affective Computing}} \bibinfo{volume}{3}, \bibinfo{number}{1} (\bibinfo{year}{2011}), \bibinfo{pages}{18--31}.
\newblock


\bibitem[Ko{\l}akowska et~al\mbox{.}(2014)]%
        {kolakowska2014emotion}
\bibfield{author}{\bibinfo{person}{Agata Ko{\l}akowska}, \bibinfo{person}{Agnieszka Landowska}, \bibinfo{person}{Mariusz Szwoch}, \bibinfo{person}{Wioleta Szwoch}, {and} \bibinfo{person}{Michal~R Wrobel}.} \bibinfo{year}{2014}\natexlab{}.
\newblock \showarticletitle{Emotion recognition and its applications}.
\newblock \bibinfo{journal}{\emph{Human-computer Systems Interaction: Backgrounds and Applications 3}} (\bibinfo{year}{2014}), \bibinfo{pages}{51--62}.
\newblock


\bibitem[Kossaifi et~al\mbox{.}(2019)]%
        {kossaifi2019sewa}
\bibfield{author}{\bibinfo{person}{Jean Kossaifi}, \bibinfo{person}{Robert Walecki}, \bibinfo{person}{Yannis Panagakis}, \bibinfo{person}{Jie Shen}, \bibinfo{person}{Maximilian Schmitt}, \bibinfo{person}{Fabien Ringeval}, \bibinfo{person}{Jing Han}, \bibinfo{person}{Vedhas Pandit}, \bibinfo{person}{Antoine Toisoul}, \bibinfo{person}{Bj{\"o}rn Schuller}, {et~al\mbox{.}}} \bibinfo{year}{2019}\natexlab{}.
\newblock \showarticletitle{Sewa db: A rich database for audio-visual emotion and sentiment research in the wild}.
\newblock \bibinfo{journal}{\emph{IEEE Transactions on Pattern Analysis and Machine Intelligence}} \bibinfo{volume}{43}, \bibinfo{number}{3} (\bibinfo{year}{2019}), \bibinfo{pages}{1022--1040}.
\newblock


\bibitem[Li and Deng(2020)]%
        {li2020deep}
\bibfield{author}{\bibinfo{person}{Shan Li} {and} \bibinfo{person}{Weihong Deng}.} \bibinfo{year}{2020}\natexlab{}.
\newblock \showarticletitle{Deep facial expression recognition: A survey}.
\newblock \bibinfo{journal}{\emph{IEEE Transactions on Affective Computing}} \bibinfo{volume}{13}, \bibinfo{number}{3} (\bibinfo{year}{2020}), \bibinfo{pages}{1195--1215}.
\newblock


\bibitem[Li et~al\mbox{.}(2022)]%
        {li2022eeg}
\bibfield{author}{\bibinfo{person}{Xiang Li}, \bibinfo{person}{Yazhou Zhang}, \bibinfo{person}{Prayag Tiwari}, \bibinfo{person}{Dawei Song}, \bibinfo{person}{Bin Hu}, \bibinfo{person}{Meihong Yang}, \bibinfo{person}{Zhigang Zhao}, \bibinfo{person}{Neeraj Kumar}, {and} \bibinfo{person}{Pekka Marttinen}.} \bibinfo{year}{2022}\natexlab{}.
\newblock \showarticletitle{EEG based emotion recognition: A tutorial and review}.
\newblock \bibinfo{journal}{\emph{Comput. Surveys}} \bibinfo{volume}{55}, \bibinfo{number}{4} (\bibinfo{year}{2022}), \bibinfo{pages}{1--57}.
\newblock


\bibitem[Lin et~al\mbox{.}(2019)]%
        {lin2019tsm}
\bibfield{author}{\bibinfo{person}{Ji Lin}, \bibinfo{person}{Chuang Gan}, {and} \bibinfo{person}{Song Han}.} \bibinfo{year}{2019}\natexlab{}.
\newblock \showarticletitle{Tsm: Temporal shift module for efficient video understanding}. In \bibinfo{booktitle}{\emph{IEEE/CVF International Conference on Computer Vision}}. \bibinfo{pages}{7083--7093}.
\newblock


\bibitem[Liu et~al\mbox{.}(2016)]%
        {liu2016retracted}
\bibfield{author}{\bibinfo{person}{Mingyang Liu}, \bibinfo{person}{Di Fan}, \bibinfo{person}{Xiaohan Zhang}, {and} \bibinfo{person}{Xiaopeng Gong}.} \bibinfo{year}{2016}\natexlab{}.
\newblock \showarticletitle{Retracted: Human emotion recognition based on galvanic skin response signal feature selection and svm}. In \bibinfo{booktitle}{\emph{International Conference on Smart City and Systems Engineering}}. IEEE, \bibinfo{pages}{157--160}.
\newblock


\bibitem[Liu et~al\mbox{.}(2021)]%
        {liu2021imigue}
\bibfield{author}{\bibinfo{person}{Xin Liu}, \bibinfo{person}{Henglin Shi}, \bibinfo{person}{Haoyu Chen}, \bibinfo{person}{Zitong Yu}, \bibinfo{person}{Xiaobai Li}, {and} \bibinfo{person}{Guoying Zhao}.} \bibinfo{year}{2021}\natexlab{}.
\newblock \showarticletitle{iMiGUE: An identity-free video dataset for micro-gesture understanding and emotion analysis}. In \bibinfo{booktitle}{\emph{IEEE/CVF Conference on Computer Vision and Pattern Recognition}}. \bibinfo{pages}{10631--10642}.
\newblock


\bibitem[Liu et~al\mbox{.}(2022)]%
        {liu2022video}
\bibfield{author}{\bibinfo{person}{Ze Liu}, \bibinfo{person}{Jia Ning}, \bibinfo{person}{Yue Cao}, \bibinfo{person}{Yixuan Wei}, \bibinfo{person}{Zheng Zhang}, \bibinfo{person}{Stephen Lin}, {and} \bibinfo{person}{Han Hu}.} \bibinfo{year}{2022}\natexlab{}.
\newblock \showarticletitle{Video swin transformer}. In \bibinfo{booktitle}{\emph{IEEE/CVF Conference on Computer Vision and Pattern Recognition}}. \bibinfo{pages}{3202--3211}.
\newblock


\bibitem[Loi et~al\mbox{.}(2013)]%
        {loi2013recognition}
\bibfield{author}{\bibinfo{person}{Felice Loi}, \bibinfo{person}{Jatin~G Vaidya}, {and} \bibinfo{person}{Sergio Paradiso}.} \bibinfo{year}{2013}\natexlab{}.
\newblock \showarticletitle{Recognition of emotion from body language among patients with unipolar depression}.
\newblock \bibinfo{journal}{\emph{Psychiatry research}} \bibinfo{volume}{209}, \bibinfo{number}{1} (\bibinfo{year}{2013}), \bibinfo{pages}{40--49}.
\newblock


\bibitem[McAdams(1984)]%
        {mcadams1984spectral}
\bibfield{author}{\bibinfo{person}{Stephen~Edward McAdams}.} \bibinfo{year}{1984}\natexlab{}.
\newblock \bibinfo{booktitle}{\emph{Spectral fusion, spectral parsing and the formation of auditory images}}.
\newblock \bibinfo{publisher}{Stanford university}.
\newblock


\bibitem[McDuff et~al\mbox{.}(2013)]%
        {mcduff2013affectiva}
\bibfield{author}{\bibinfo{person}{Daniel McDuff}, \bibinfo{person}{Rana Kaliouby}, \bibinfo{person}{Thibaud Senechal}, \bibinfo{person}{May Amr}, \bibinfo{person}{Jeffrey Cohn}, {and} \bibinfo{person}{Rosalind Picard}.} \bibinfo{year}{2013}\natexlab{}.
\newblock \showarticletitle{Affectiva-mit facial expression dataset (am-fed): Naturalistic and spontaneous facial expressions collected}. In \bibinfo{booktitle}{\emph{IEEE Conference on Computer Vision and Pattern Recognition Workshops}}. \bibinfo{pages}{881--888}.
\newblock


\bibitem[Mollahosseini et~al\mbox{.}(2017)]%
        {mollahosseini2017affectnet}
\bibfield{author}{\bibinfo{person}{Ali Mollahosseini}, \bibinfo{person}{Behzad Hasani}, {and} \bibinfo{person}{Mohammad~H Mahoor}.} \bibinfo{year}{2017}\natexlab{}.
\newblock \showarticletitle{Affectnet: A database for facial expression, valence, and arousal computing in the wild}.
\newblock \bibinfo{journal}{\emph{IEEE Transactions on Affective Computing}} \bibinfo{volume}{10}, \bibinfo{number}{1} (\bibinfo{year}{2017}), \bibinfo{pages}{18--31}.
\newblock


\bibitem[Morency et~al\mbox{.}(2011)]%
        {morency2011towards}
\bibfield{author}{\bibinfo{person}{Louis-Philippe Morency}, \bibinfo{person}{Rada Mihalcea}, {and} \bibinfo{person}{Payal Doshi}.} \bibinfo{year}{2011}\natexlab{}.
\newblock \showarticletitle{Towards multimodal sentiment analysis: Harvesting opinions from the web}. In \bibinfo{booktitle}{\emph{International Conference on Multimodal Interfaces}}. \bibinfo{pages}{169--176}.
\newblock


\bibitem[Nandwani and Verma(2021)]%
        {nandwani2021review}
\bibfield{author}{\bibinfo{person}{Pansy Nandwani} {and} \bibinfo{person}{Rupali Verma}.} \bibinfo{year}{2021}\natexlab{}.
\newblock \showarticletitle{A review on sentiment analysis and emotion detection from text}.
\newblock \bibinfo{journal}{\emph{Social network analysis and mining}} \bibinfo{volume}{11}, \bibinfo{number}{1} (\bibinfo{year}{2021}), \bibinfo{pages}{81}.
\newblock


\bibitem[Navarro and Karlins(2008)]%
        {navarro2008every}
\bibfield{author}{\bibinfo{person}{Julia Navarro} {and} \bibinfo{person}{Marvin Karlins}.} \bibinfo{year}{2008}\natexlab{}.
\newblock \bibinfo{booktitle}{\emph{What every body is saying}}.
\newblock \bibinfo{publisher}{HarperCollins Publishers New York, NY, USA:}.
\newblock


\bibitem[Ouyang et~al\mbox{.}(2022)]%
        {ouyang2022training}
\bibfield{author}{\bibinfo{person}{Long Ouyang}, \bibinfo{person}{Jeffrey Wu}, \bibinfo{person}{Xu Jiang}, \bibinfo{person}{Diogo Almeida}, \bibinfo{person}{Carroll Wainwright}, \bibinfo{person}{Pamela Mishkin}, \bibinfo{person}{Chong Zhang}, \bibinfo{person}{Sandhini Agarwal}, \bibinfo{person}{Katarina Slama}, \bibinfo{person}{Alex Ray}, {et~al\mbox{.}}} \bibinfo{year}{2022}\natexlab{}.
\newblock \showarticletitle{Training language models to follow instructions with human feedback}.
\newblock \bibinfo{journal}{\emph{Advances in neural information processing systems}}  \bibinfo{volume}{35} (\bibinfo{year}{2022}), \bibinfo{pages}{27730--27744}.
\newblock


\bibitem[Patino et~al\mbox{.}(2020)]%
        {patino2020speaker}
\bibfield{author}{\bibinfo{person}{Jose Patino}, \bibinfo{person}{Natalia Tomashenko}, \bibinfo{person}{Massimiliano Todisco}, \bibinfo{person}{Andreas Nautsch}, {and} \bibinfo{person}{Nicholas Evans}.} \bibinfo{year}{2020}\natexlab{}.
\newblock \showarticletitle{Speaker anonymisation using the McAdams coefficient}.
\newblock \bibinfo{journal}{\emph{arXiv preprint arXiv:2011.01130}} (\bibinfo{year}{2020}).
\newblock


\bibitem[Rahdari et~al\mbox{.}(2019)]%
        {rahdari2019multimodal}
\bibfield{author}{\bibinfo{person}{Farhad Rahdari}, \bibinfo{person}{Esmat Rashedi}, {and} \bibinfo{person}{Mahdi Eftekhari}.} \bibinfo{year}{2019}\natexlab{}.
\newblock \showarticletitle{A multimodal emotion recognition system using facial landmark analysis}.
\newblock \bibinfo{journal}{\emph{Iranian Journal of Science and Technology, Transactions of Electrical Engineering}}  \bibinfo{volume}{43} (\bibinfo{year}{2019}), \bibinfo{pages}{171--189}.
\newblock


\bibitem[Tomashenko et~al\mbox{.}(2024)]%
        {tomashenko2024voiceprivacy}
\bibfield{author}{\bibinfo{person}{Natalia Tomashenko}, \bibinfo{person}{Xiaoxiao Miao}, \bibinfo{person}{Pierre Champion}, \bibinfo{person}{Sarina Meyer}, \bibinfo{person}{Xin Wang}, \bibinfo{person}{Emmanuel Vincent}, \bibinfo{person}{Michele Panariello}, \bibinfo{person}{Nicholas Evans}, \bibinfo{person}{Junichi Yamagishi}, {and} \bibinfo{person}{Massimiliano Todisco}.} \bibinfo{year}{2024}\natexlab{}.
\newblock \showarticletitle{The {VoicePrivacy} 2024 Challenge Evaluation Plan}.
\newblock  (\bibinfo{year}{2024}).
\newblock
\showeprint[arxiv]{2404.02677}~[eess.AS]


\bibitem[Touvron et~al\mbox{.}(2023)]%
        {touvron2023llama}
\bibfield{author}{\bibinfo{person}{Hugo Touvron}, \bibinfo{person}{Louis Martin}, \bibinfo{person}{Kevin Stone}, \bibinfo{person}{Peter Albert}, \bibinfo{person}{Amjad Almahairi}, \bibinfo{person}{Yasmine Babaei}, \bibinfo{person}{Nikolay Bashlykov}, \bibinfo{person}{Soumya Batra}, \bibinfo{person}{Prajjwal Bhargava}, \bibinfo{person}{Shruti Bhosale}, {et~al\mbox{.}}} \bibinfo{year}{2023}\natexlab{}.
\newblock \showarticletitle{Llama 2: Open foundation and fine-tuned chat models}.
\newblock \bibinfo{journal}{\emph{arXiv preprint arXiv:2307.09288}} (\bibinfo{year}{2023}).
\newblock


\bibitem[Zadeh et~al\mbox{.}(2018)]%
        {zadeh2018multimodal}
\bibfield{author}{\bibinfo{person}{AmirAli~Bagher Zadeh}, \bibinfo{person}{Paul~Pu Liang}, \bibinfo{person}{Soujanya Poria}, \bibinfo{person}{Erik Cambria}, {and} \bibinfo{person}{Louis-Philippe Morency}.} \bibinfo{year}{2018}\natexlab{}.
\newblock \showarticletitle{Multimodal language analysis in the wild: Cmu-mosei dataset and interpretable dynamic fusion graph}. In \bibinfo{booktitle}{\emph{Annual Meeting of the Association for Computational Linguistics}}. \bibinfo{pages}{2236--2246}.
\newblock


\bibitem[Zhang et~al\mbox{.}(2023)]%
        {damonlpsg2023videollama}
\bibfield{author}{\bibinfo{person}{Hang Zhang}, \bibinfo{person}{Xin Li}, {and} \bibinfo{person}{Lidong Bing}.} \bibinfo{year}{2023}\natexlab{}.
\newblock \showarticletitle{Video-LLaMA: An Instruction-tuned Audio-Visual Language Model for Video Understanding}.
\newblock \bibinfo{journal}{\emph{arXiv preprint arXiv:2306.02858}} (\bibinfo{year}{2023}).
\newblock
\urldef\tempurl%
\url{https://arxiv.org/abs/2306.02858}
\showURL{%
\tempurl}


\bibitem[Zhang et~al\mbox{.}(2024)]%
        {zhang2024vision}
\bibfield{author}{\bibinfo{person}{Jingyi Zhang}, \bibinfo{person}{Jiaxing Huang}, \bibinfo{person}{Sheng Jin}, {and} \bibinfo{person}{Shijian Lu}.} \bibinfo{year}{2024}\natexlab{}.
\newblock \showarticletitle{Vision-language models for vision tasks: A survey}.
\newblock \bibinfo{journal}{\emph{IEEE Transactions on Pattern Analysis and Machine Intelligence}} (\bibinfo{year}{2024}).
\newblock


\end{thebibliography}

\end{document}